\PassOptionsToPackage{unicode}{hyperref}
\PassOptionsToPackage{hyphens}{url}
\PassOptionsToPackage{dvipsnames,svgnames,x11names}{xcolor}
\documentclass[
  11pt,
]{article}
\usepackage{xcolor}
\usepackage[a4paper,twoside,inner=2.5cm,outer=2.0cm,top=2.5cm,bottom=2.5cm]{geometry}
\usepackage{amsmath,amssymb}
\usepackage{iftex}
\ifPDFTeX
  \usepackage[T1]{fontenc}
  \usepackage[utf8]{inputenc}
  \usepackage{textcomp} % provide euro and other symbols
\else % if luatex or xetex
  \usepackage{unicode-math} % this also loads fontspec
  \defaultfontfeatures{Scale=MatchLowercase}
  \defaultfontfeatures[\rmfamily]{Ligatures=TeX,Scale=1}
\fi
\usepackage{lmodern}
\ifPDFTeX\else
\fi
\IfFileExists{upquote.sty}{\usepackage{upquote}}{}
\IfFileExists{microtype.sty}{% use microtype if available
  \usepackage[]{microtype}
  \UseMicrotypeSet[protrusion]{basicmath} % disable protrusion for tt fonts
}{}
\makeatletter
\@ifundefined{KOMAClassName}{% if non-KOMA class
  \IfFileExists{parskip.sty}{%
    \usepackage{parskip}
  }{% else
    \setlength{\parindent}{0pt}
    \setlength{\parskip}{6pt plus 2pt minus 1pt}}
}{% if KOMA class
  \KOMAoptions{parskip=half}}
\makeatother
\usepackage{color}
\usepackage{fancyvrb}

\DefineVerbatimEnvironment{Highlighting}{Verbatim}{commandchars=\\\{\}}
\newenvironment{Shaded}{}{}

\newcommand{\AttributeTok}[1]{\textcolor[rgb]{0.49,0.56,0.16}{#1}}

\newcommand{\BuiltInTok}[1]{\textcolor[rgb]{0.00,0.50,0.00}{#1}}

\newcommand{\CommentTok}[1]{\textcolor[rgb]{0.38,0.63,0.69}{\textit{#1}}}

\newcommand{\ExtensionTok}[1]{#1}

\newcommand{\FunctionTok}[1]{\textcolor[rgb]{0.02,0.16,0.49}{#1}}

\newcommand{\NormalTok}[1]{#1}
\newcommand{\OperatorTok}[1]{\textcolor[rgb]{0.40,0.40,0.40}{#1}}

\newcommand{\StringTok}[1]{\textcolor[rgb]{0.25,0.44,0.63}{#1}}
\newcommand{\VariableTok}[1]{\textcolor[rgb]{0.10,0.09,0.49}{#1}}

\usepackage{longtable,booktabs,array}
\usepackage{calc} % for calculating minipage widths
\usepackage{etoolbox}
\makeatletter
\patchcmd\longtable{\par}{\if@noskipsec\mbox{}\fi\par}{}{}
\makeatother
\IfFileExists{footnotehyper.sty}{\usepackage{footnotehyper}}{\usepackage{footnote}}
\makesavenoteenv{longtable}
\usepackage{graphicx}
\makeatletter
\newsavebox\pandoc@box
\newcommand*\pandocbounded[1]{% scales image to fit in text height/width
  \sbox\pandoc@box{#1}%
  \Gscale@div\@tempa{\textheight}{\dimexpr\ht\pandoc@box+\dp\pandoc@box\relax}%
  \Gscale@div\@tempb{\linewidth}{\wd\pandoc@box}%
  \ifdim\@tempb\p@<\@tempa\p@\let\@tempa\@tempb\fi% select the smaller of both
  \ifdim\@tempa\p@<\p@\scalebox{\@tempa}{\usebox\pandoc@box}%
  \else\usebox{\pandoc@box}%
  \fi%
}
\def\fps@figure{htbp}
\makeatother
\providecommand{\tightlist}{%
  \setlength{\itemsep}{0pt}\setlength{\parskip}{0pt}}

\usepackage{pdflscape}
\usepackage{booktabs}
\usepackage{caption}
\usepackage{eso-pic}
\AddToShipoutPictureBG{%
  \ifdim\pdfpagewidth>\pdfpageheight\relax
    \put(24,298){%
      \rotatebox{90}{\normalfont\normalsize\thepage}%
    }%
  \fi
}

\makeatletter
\renewcommand{\@maketitle}{%
  \newpage\null\vskip 1em%
  \begin{center}%
    \let\footnote\thanks%
    {\Large\bfseries\@title\par}%
    \vskip 0.6em%
    {\small\begin{tabular}[t]{c}\@author\end{tabular}\par}%
    \vskip 0.3em%
    {\small\itshape\@date}%
  \end{center}%
  \par\vskip 1em%
}
\makeatother
\usepackage{bookmark}
\IfFileExists{xurl.sty}{\usepackage{xurl}}{} % add URL line breaks if available
\makeatletter
\@ifundefined{xmpquote}{}{}
\makeatother
\hypersetup{
  pdftitle={The African Language Tax},
  pdfauthor={Olaoye Anthony Somide, DataLens Africa Research · CipherSense AI Technologies Ltd.},
  colorlinks=true,
  linkcolor={blue},
  filecolor={Maroon},
  citecolor={Blue},
  urlcolor={blue},
  pdfcreator={LaTeX via pandoc}}

\title{The African Language Tax}
\usepackage{etoolbox}
\makeatletter
\providecommand{\subtitle}[1]{% add subtitle to \maketitle
  \apptocmd{\@title}{\par {\large #1 \par}}{}{}
}
\makeatother
\subtitle{Quantifying the Cost, Latency, and Context Penalty of
Tokenizing African Languages in Frontier LLMs}
\author{Olaoye Anthony Somide, DataLens Africa Research · CipherSense AI
Technologies Ltd.\thanks{Correspondence: olaoye.somide@ciphersense.ai}}
\date{June 2026}

\begin{document}
\maketitle

\subsection*{Abstract}\label{abstract}

Commercial large language models bill, scale latency, and budget context
per token. Yet tokenizers assign more subword tokens to the same meaning
in some languages than in others, so speakers of languages with high
token-fertility pay a structural penalty before a model is ever invoked.
This penalty is documented for multilingual settings in general, but it
has not been measured systematically for African languages at the level
of enterprise deployment economics and cognitive context capacity. We
measure it across 20 African languages spanning five language families
and three scripts (Latin, Ge'ez/Ethiopic, N'Ko; 19 appear in the primary
FLORES-200+ corpus, with Nigerian Pidgin measured via MAFAND-MT only),
using parallel corpora so that the language effect is isolated from
content. Across 11 frontier and open tokenizers on FLORES-200+, every
African language carries a tokenization premium above English (median
1.88× on GPT-5 / o200k\_base, up to 8.92× for N'Ko); the penalty is
largest for Ethiopic and N'Ko scripts (reaching 7--9×) and is
near-invariant across corpora (FLORES vs SIB-200 Pearson r = 0.9998).
Translated into deployment terms, this results in up to 8.9× inference
cost and an equivalent generation-latency multiplier (N'Ko vs English on
GPT-5; 7.4× for Amharic), and as little as 11\% of English's effective
context window. The best currently available tokenizer for African
languages, Gemma 4, reduces the mean premium from 3.31× (cl100k\_base)
to 2.38×, but no tokenizer eliminates the penalty. We release an open
measurement tool (\texttt{afri-fertility}), a public leaderboard, a
results dataset, and mitigation guidance for African builders. The
penalty falls hardest on the languages whose speakers can least afford
it, a digital divide encoded directly into the subword vocabulary.

\subsection*{Keywords}\label{keywords}

tokenization, subword fertility, African languages, LLM inference cost,
multilingual NLP, tokenizer fairness, low-resource languages, AI equity

\section{1. Introduction}\label{introduction}

\subsection{1.1 The Pre-Inference Cost
Layer}\label{the-pre-inference-cost-layer}

Before a large language model reasons about a single word of a prompt,
it must first convert that text into tokens. Commercial LLM providers
price their APIs per token, scale generation latency with the number of
tokens produced, and bound the conversation by a fixed token context
window. Tokenization is therefore not a neutral preprocessing step: it
is the layer at which the economics of using a model are set. Two
prompts that carry exactly the same meaning, but that tokenize into
different numbers of tokens, cost different amounts to process, take
different lengths of time to answer, and consume different fractions of
the available context (effectively reducing the model's ``operational
memory'' and forcing earlier truncation of conversation history) before
the model's quality, accuracy, or capability enters the picture at all.

This matters because tokenizers do not segment all languages equally. A
subword vocabulary learned predominantly from English and other
high-resource, web-abundant languages represents those languages
compactly, while fragmenting others into many short pieces. The standard
measure of this effect is \emph{fertility} (the number of subword tokens
a tokenizer emits per word) and the disparity across languages is large
and well documented. \hyperref[ref:petrov2023]{Petrov et al.~2023} show
that the same content can require up to roughly fifteen times as many
tokens in one language as in another under current tokenizers, and
\hyperref[ref:ahia2023]{Ahia et al.~2023} demonstrate that this length
disparity translates directly into unequal API cost and utility across
typologically diverse languages. The penalty is structural: it is fixed
by the tokenizer's vocabulary, applies on every request, and cannot be
engineered away by the downstream user.

\subsection{1.2 Why African Languages, and Why
Now}\label{why-african-languages-and-why-now}

The cross-lingual tokenization gap has been documented in general
multilingual settings and, most directly, for European languages by
\hyperref[ref:ovcharov2026]{Ovcharov 2026}, who measures the ``tokenizer
tax'' across 25 languages on parallel text, finding that English encodes
at roughly 1.2 tokens per word while Greek and Maltese reach about 3.1×,
with efficiency rankings near-invariant across text registers. That
study is the closest methodological precedent to ours, but its largest
penalties fall on European languages that still share the Latin or
Greek/Cyrillic script families and substantial web presence. African
languages have been left unmeasured at this resolution, even though they
combine exactly the properties that should push the penalty well past
that ceiling: non-Latin scripts (Geʿez/Ethiopic, Arabic/Ajami, N'Ko),
agglutinative and tonal morphology, and far thinner representation in
tokenizer training corpora. The gap is consequential: African builders
deploying customer-service, clinical-triage, and agricultural advisory
services must increasingly operate in local languages at production
scale, in markets where compute is least affordable; and the African NLP
community has invested heavily in accuracy benchmarks (IrokoBench
\hyperref[ref:adelani2025iroko]{Adelani et al.~2025}, AfroBench
\hyperref[ref:ojo2023]{Ojo et al.~2023}) while the pre-inference cost
layer beneath them has gone unquantified. A model can score well on a
task and still be uneconomic to run in the language it is written in.

We address this with a parallel-corpus measurement. Because parallel
corpora express the same meaning across every language, differences in
token counts for a fixed tokenizer reflect the language and its script,
not the content, isolating exactly the effect we want to measure. We
translate these results into the terms a decision-maker uses: USD and
local currency costs, latency multipliers, and the effective capacity of
the model's context window.

\subsection{1.3 Contributions}\label{contributions}

This paper makes three contributions:

\begin{enumerate}
\def\labelenumi{\arabic{enumi}.}
\item
  \textbf{The first comprehensive audit of the African-language
  tokenization premium}, establishing a baseline for digital equity
  across 20 African languages and 11 current frontier and open
  tokenizers. We report fertility and an English-relative premium for
  each language--tokenizer pair across multiple script families, with
  corpus-level aggregation and bootstrap confidence intervals.
\item
  \textbf{An enterprise cost model} that converts fertility into USD and
  local-currency (NGN, ZAR, KES) cost, accounting for both
  token-fertility and the compounding effect of local currency
  volatility against USD-denominated API pricing. We further quantify
  generation-latency multipliers and context-window erosion, which
  reduces the ``operational memory'' of the model and forces
  African-language applications to handle shorter conversation histories
  for the same cost. We instantiate this model on three concrete
  deployment scenarios: a bank customer-service assistant, a
  clinical-triage line, and an SMS agricultural-advisory service.
\item
  \textbf{Open artifacts}:
  \href{https://github.com/CipherSenseAI/afri-fertility}{\texttt{afri-fertility}},
  a deterministic measurement tool with an African test suite built in;
  a public African Tokenization Tax Leaderboard; a released results
  dataset; and a one-page mitigation guide for African builders.
\end{enumerate}

Together these turn a known but abstract inequality into a measured,
reproducible, and economically legible penalty for the African
deployment context, with tooling for anyone to re-run and extend the
measurement.

\section{2. Background \& Related Work}\label{background-related-work}

\subsection{2.1 Subword Tokenization and
Fertility}\label{subword-tokenization-and-fertility}

Modern LLMs operate over subword vocabularies learned from a training
corpus, typically by byte-pair encoding (BPE), unigram/SentencePiece, or
a byte-level BPE variant such as the \texttt{tiktoken} family used by
recent OpenAI models. The vocabulary fixes how any string is split:
sequences frequent in the training corpus are merged into single tokens,
while rarer sequences are left as many short pieces, down to individual
bytes for scripts the vocabulary barely covers. The standard scalar
summary of this behaviour is \emph{fertility} (tokens emitted per word
\hyperref[ref:rust2021]{Rust et al.~2021}), which Rust et al.~link to
downstream model quality and which we adopt here as the basis for an
English-relative \emph{premium}. Because a vocabulary is a frozen
artifact of its training mixture, its fertility on a given language is
fixed at training time and borne by every subsequent request; a
deploying user cannot reduce it without changing the model.

\subsection{2.2 Cross-Lingual Tokenization
Inequality}\label{cross-lingual-tokenization-inequality}

That fertility varies sharply across languages, and that the variation
is systematically tied to a language's representation in training data,
is now well established. \hyperref[ref:petrov2023]{Petrov et al.~2023}
document that the same meaning can require up to roughly fifteen times
as many tokens in one language as in another across contemporary
tokenizers, and frame this as an unfairness baked into the tokenizer
itself. \hyperref[ref:ahia2023]{Ahia et al.~2023} connect the length
disparity to deployment directly: under per-token API pricing, speakers
of high-fertility languages pay more for the same service and exhaust
fixed context windows faster, across 22 typologically diverse languages.
These two results establish the phenomenon and its economic consequence
in general multilingual terms. Neither isolates the African case at
depth, and both predate the current generation of frontier tokenizers we
measure.

\subsection{2.3 Recent ``Token Tax''
Framings}\label{recent-token-tax-framings}

Two recent works name the effect as a ``tax.''
\hyperref[ref:lundin2026]{Lundin et al.~2026} (\emph{The Token Tax})
characterizes systematic bias in multilingual tokenization in general
terms. Closest to our method, \hyperref[ref:ovcharov2026]{Ovcharov 2026}
measures the tokenizer tax across 25 European languages and ten
foundation models on parallel text, reporting fertility from
\textasciitilde1.2 tokens/word for English to \textasciitilde3.1 for
Greek and Maltese (a \textasciitilde2.5× penalty), near-invariance of
efficiency rankings across registers (correlation \textgreater{} 0.97),
fragmentation at morphological boundaries, and a release of all
measurements as a public dataset. Ovcharov is the methodological
template we build on: parallel-corpus fertility across many tokenizers,
openly released. The decisive difference is scope and framing. The
European ceiling is set by languages that retain Latin, Greek, or
Cyrillic scripts and meaningful web presence; African languages add
non-Latin scripts (Ge'ez/Ethiopic, Arabic/Ajami, N'Ko), heavier
agglutinative and tonal morphology, and far thinner training
representation; and no prior work translates the resulting fertility
into the enterprise cost, latency, and context terms an African deployer
must reason about, nor ships the measurement as an open tool and
leaderboard.

\subsection{2.4 African NLP Evaluation}\label{african-nlp-evaluation}

The African NLP community has built a substantial evaluation
infrastructure focused on model \emph{accuracy}. IrokoBench
\hyperref[ref:adelani2025iroko]{Adelani et al.~2025} and AfroBench
\hyperref[ref:ojo2023]{Ojo et al.~2023} benchmark frontier and open
models across many African languages and tasks, and the broader
Masakhane ecosystem (including the FLORES-200+ parallel corpus
\hyperref[ref:nllb2024]{NLLB Team 2024} we use as our primary
measurement set) supplies the parallel and labelled data the field
relies on. This work measures whether a model is \emph{correct} in a
language. It does not measure the cost layer beneath that correctness:
how many tokens, and therefore how many dollars, how much latency, and
how much context, the language consumes regardless of accuracy. The two
are complementary, and we join them directly by relating our
per-language premiums to published IrokoBench/AfroBench accuracy to test
whether higher-cost languages also tend to be lower-accuracy ones (H4).
Separately, \hyperref[ref:ndomba2025]{Ndomba et al.~2025} find that
language-specific tokenizers can outperform multilingual defaults on
African-language tasks, which motivates our mitigation analysis (§8).

\subsection{2.5 Positioning}\label{positioning}

This paper is the first to measure the tokenization penalty for African
languages across multiple families and scripts on parallel corpora and
current frontier tokenizers, and to translate it into a concrete
enterprise cost, latency, and context model, shipped as an open tool,
leaderboard, and dataset. \hyperref[tbl:2.1]{Table 2.1} positions the
contribution against the nearest prior work.

\phantomsection\label{tbl:2.1}

\textbf{Table 2.1: Related-work positioning}

{\def\LTcaptype{none} % do not increment counter
\begin{longtable}[]{@{}
  >{\raggedright\arraybackslash}p{(\linewidth - 4\tabcolsep) * \real{0.3333}}
  >{\raggedright\arraybackslash}p{(\linewidth - 4\tabcolsep) * \real{0.3333}}
  >{\raggedright\arraybackslash}p{(\linewidth - 4\tabcolsep) * \real{0.3333}}@{}}
\toprule\noalign{}
\begin{minipage}[b]{\linewidth}\raggedright
Work
\end{minipage} & \begin{minipage}[b]{\linewidth}\raggedright
What it did
\end{minipage} & \begin{minipage}[b]{\linewidth}\raggedright
How this paper differs
\end{minipage} \\
\midrule\noalign{}
\endhead
\bottomrule\noalign{}
\endlastfoot
\hyperref[ref:petrov2023]{Petrov et al.~2023} (NeurIPS) & Tokenizer
unfairness across languages; up to \textasciitilde15× length disparities
& Africa-focused depth; current frontier tokenizers; enterprise cost
model \\
\hyperref[ref:ahia2023]{Ahia et al.~2023} (EMNLP) & API cost/utility
across 22 typologically diverse languages & African families/scripts at
depth; named deployment scenarios \\
\hyperref[ref:lundin2026]{Lundin et al.~2026} (\emph{The Token Tax}) &
Multilingual tokenization bias (general) & Africa-specific;
parallel-corpus; economics; open tool + leaderboard \\
\hyperref[ref:ovcharov2026]{Ovcharov 2026} (\emph{Tokenizer Tax across
25 European Languages}) & Parallel-corpus fertility for European
languages & African languages and non-Latin scripts; enterprise
framing \\
African NLP benchmarks: IrokoBench
\hyperref[ref:adelani2025iroko]{Adelani et al.~2025}, AfroBench
\hyperref[ref:ojo2023]{Ojo et al.~2023} & Accuracy benchmarks & Measures
the \emph{pre-inference} cost layer; links to their accuracy for H4 \\
\end{longtable}
}

\section{3. Definitions \& Metrics}\label{definitions-metrics}

\subsection{3.1 Notation and Counting
Conventions}\label{notation-and-counting-conventions}

Let a \emph{document} be a piece of text in language \(L\), tokenized by
tokenizer \(T\). For a document we define four base counts, each
computed under a single fixed convention applied identically to every
language so that no language receives special preprocessing:

\begin{itemize}
\tightlist
\item
  \(W(L)\), \textbf{word count}, obtained by Unicode word segmentation
  (UAX-29, via ICU/\texttt{uniseg}). We pre-specify segmentation because
  most target languages are whitespace-delimited and the choice must be
  identical across languages to keep fertility comparable. A
  \texttt{regex\ \textbackslash{}w+} fallback is recorded in run
  metadata if the primary segmenter is unavailable.
\item
  \(N(L,T)\), \textbf{token count}: the number of subword tokens \(T\)
  produces, \textbf{excluding} special, BOS, and EOS tokens. This is
  pre-specified so that fixed per-sequence overhead does not inflate
  short-sentence counts unevenly.
\item
  \(\mathrm{chars}(L)\), \textbf{character count} in Unicode scalar
  values.
\item
  \(\mathrm{bytes}(L)\), \textbf{UTF-8 byte count}.
\end{itemize}

All text is normalized to Unicode \textbf{NFC} before any count is
taken. Empty or whitespace-only inputs yield zero counts without error.

\subsection{3.2 Metrics (Locked
Formulas)}\label{metrics-locked-formulas}

From the base counts we derive the study's five metrics. Lower fertility
(and lower premium) is better.

\textbf{Fertility}, tokens per word; the primary efficiency measure.

\[F(L,T) = \frac{N(L,T)}{W(L)}\]

\textbf{Premium (parity)}, fertility relative to the baseline language
(English by default); the headline number.

\[P(L,T) = \frac{F(L,T)}{F(\text{eng},T)}\]

\(P\) reads directly as: \emph{speakers of \(L\) pay \(P\times\) the
tokens English speakers pay for the same meaning.} By construction
\(P(\text{eng},T) = 1\) for every \(T\), and we expect \(P > 0\)
everywhere.

\textbf{Characters per token}, compression efficiency.

\[\mathrm{CPT}(L,T) = \frac{\mathrm{chars}(L)}{N(L,T)}\]

\textbf{Bytes per token}, a byte-level view that is fairer across
scripts of differing character density.

\[\mathrm{BPT}(L,T) = \frac{\mathrm{bytes}(L)}{N(L,T)}\]

\textbf{Context efficiency}, how much real content fits in a fixed
context window of size \(\text{w}\), and its value relative to the
baseline.

\[\mathrm{CE}(L,T) = \text{w} \cdot \mathrm{CPT}(L,T), \qquad \mathrm{relCE}(L,T) = \frac{\mathrm{CE}(L,T)}{\mathrm{CE}(\text{eng},T)}\]

We report \(\mathrm{relCE}\) at the pre-specified window size
\(\text{w} = 128{,}000\).

\subsection{3.3 Why Parallel Corpora}\label{why-parallel-corpora}

Every corpus in this study is \textbf{parallel}: each item expresses the
same meaning across all languages. This is the design's central control.
For a fixed tokenizer \(T\), holding meaning constant means that
differences in \(F(L,T)\) across languages cannot be attributed to
differences in \emph{what is being said}, only to \emph{how the language
and its script encode it}. Fertility measured on non-parallel corpora
confounds the language effect with topic, length, and register; the
parallel design removes that confound, which is what licenses reading
\(P(L,T)\) as a property of the language--tokenizer pair rather than of
the sample.

\subsection{3.4 Aggregation and
Uncertainty}\label{aggregation-and-uncertainty}

Corpus-level metrics are computed by \textbf{sum-then-divide}, not
mean-of-ratios. For a corpus of sentences indexed by \(i\):

\[F(L,T) = \frac{\sum_i N_i(L,T)}{\sum_i W_i(L)}\]

and analogously for CPT and BPT, with \(P\) formed from the aggregated
fertilities. Sum-then-divide weights each sentence by its length and
avoids the instability and bias that mean-of-ratios introduces on short
sentences; the two are not interchangeable, and the tool explicitly
guards against conflating them.

We attach bootstrap 95\% confidence intervals to fertility and premium
(1,000 resamples over sentences) to show that estimates are stable
across different draws from the corpus. Because tokenization is
deterministic, these intervals reflect sampling variation over text, not
measurement noise.

UAX-29 word segmentation is imperfect for highly agglutinative languages
(e.g., Kinyarwanda, isiXhosa) and for Ethiopic script, where word
boundaries do not align cleanly with the whitespace/UAX-29 model.
Because \(F\) depends on \(W\), this can distort word-normalized
fertility. We therefore report the character- and byte-normalized
metrics (CPT, BPT) alongside fertility for every language, so that
conclusions never rest on word counts alone, and we flag the affected
languages in output metadata. This is revisited as a threat to validity
in §9.

\section{4. Data}\label{data}

\subsection{4.1 Languages}\label{languages}

The study covers a language set fixed in advance to span families,
scripts, and degrees of web representation, with two baselines and two
dual-script contrasts. The set is organized into tiers: a core deep-dive
(6), Latin-script breadth (12), non-Latin scripts (3), and baselines
(2). FLORES uses script-suffixed codes (e.g., \texttt{yor\_Latn},
\texttt{amh\_Ethi}).

\phantomsection\label{tbl:4.1}

\textbf{Table 4.1: Study language set, tier, family, and script}

{\def\LTcaptype{none} % do not increment counter
\begin{longtable}[]{@{}
  >{\raggedright\arraybackslash}p{(\linewidth - 10\tabcolsep) * \real{0.1667}}
  >{\raggedright\arraybackslash}p{(\linewidth - 10\tabcolsep) * \real{0.1667}}
  >{\raggedright\arraybackslash}p{(\linewidth - 10\tabcolsep) * \real{0.1667}}
  >{\raggedright\arraybackslash}p{(\linewidth - 10\tabcolsep) * \real{0.1667}}
  >{\raggedright\arraybackslash}p{(\linewidth - 10\tabcolsep) * \real{0.1667}}
  >{\raggedright\arraybackslash}p{(\linewidth - 10\tabcolsep) * \real{0.1667}}@{}}
\toprule\noalign{}
\begin{minipage}[b]{\linewidth}\raggedright
Language
\end{minipage} & \begin{minipage}[b]{\linewidth}\raggedright
ISO 639-3
\end{minipage} & \begin{minipage}[b]{\linewidth}\raggedright
Family
\end{minipage} & \begin{minipage}[b]{\linewidth}\raggedright
Script
\end{minipage} & \begin{minipage}[b]{\linewidth}\raggedright
Tier
\end{minipage} & \begin{minipage}[b]{\linewidth}\raggedright
Rationale
\end{minipage} \\
\midrule\noalign{}
\endhead
\bottomrule\noalign{}
\endlastfoot
Yoruba & yor & Niger-Congo (Volta-Niger) & Latin (tonal diacritics) &
Core & DataLens annotation language \\
Hausa & hau & Afro-Asiatic (Chadic) & Latin (+ Ajami) & Core &
Annotation pipeline; dual-script \\
Igbo & ibo & Niger-Congo (Volta-Niger) & Latin (diacritics) & Core &
Annotation pipeline \\
Wolof & wol & Niger-Congo (Atlantic) & Latin (+ Ajami) & Core &
Annotation language; dual-script \\
Swahili & swh & Niger-Congo (Bantu) & Latin & Core & Continental lingua
franca \\
Amharic & amh & Afro-Asiatic (Semitic) & Ethiopic (Ge'ez) & Core &
Extreme non-Latin case \\
Zulu & zul & Bantu & Latin & Breadth & Morphologically rich \\
Xhosa & xho & Bantu & Latin & Breadth & Click language, agglutinative \\
Shona & sna & Bantu & Latin & Breadth & \\
Kinyarwanda & kin & Bantu & Latin & Breadth & Highly agglutinative \\
Luganda & lug & Bantu & Latin & Breadth & \\
Akan/Twi & aka & Niger-Congo (Kwa) & Latin & Breadth & \\
Lingala & lin & Bantu & Latin & Breadth & Central Africa \\
Oromo & gaz & Afro-Asiatic (Cushitic) & Latin & Breadth & Large speaker
base \\
Nigerian Pidgin & pcm & English Creole & Latin & Breadth & Expected low
premium; creole effect \\
Sesotho & sot & Bantu & Latin & Breadth & \\
Bambara & bam & Mande & Latin (+ N'Ko) & Breadth & Dual-script bridge \\
Afrikaans & afr & Indo-European (Germanic) & Latin & Control & Predicted
lowest premium \\
Tigrinya & tir & Afro-Asiatic (Semitic) & Ethiopic & Non-Latin & Second
Ge'ez data point \\
Hausa (Ajami) & hau & Chadic & Arabic & Non-Latin & Script-only contrast
vs Hausa-Latin \\
Bambara (N'Ko) & bam & Mande & N'Ko & Non-Latin & Expected extreme
premium \\
English & eng & Germanic & Latin & Baseline & Premium denominator \\
French & fra & Romance & Latin & Baseline & Francophone-Africa
deployment baseline \\
\end{longtable}
}

The two dual-script entries are the design's cleanest test of the script
hypothesis (H2): Hausa appears in both Latin and Ajami (Arabic) script,
and Bambara in both Latin and N'Ko, so the script effect can be read
with the language (its morphology, vocabulary, and the parallel content)
held constant. Afrikaans is the control for H5: Latin-script, Germanic,
and web-abundant, it should show the lowest African premium if the
penalty tracks representation rather than geography. English is the
premium denominator; French is included as the Francophone-Africa
deployment baseline.

\subsection{4.2 General Parallel
Corpora}\label{general-parallel-corpora}

Three existing parallel corpora supply the general-text measurement and
two robustness checks. All are sentence-aligned, so the parallel-corpus
control of §3.3 holds across every language a corpus covers.

\phantomsection\label{tbl:4.2}

\textbf{Table 4.2: Parallel corpora used in the study}

{\def\LTcaptype{none} % do not increment counter
\begin{longtable}[]{@{}
  >{\raggedright\arraybackslash}p{(\linewidth - 10\tabcolsep) * \real{0.1667}}
  >{\raggedright\arraybackslash}p{(\linewidth - 10\tabcolsep) * \real{0.1667}}
  >{\raggedright\arraybackslash}p{(\linewidth - 10\tabcolsep) * \real{0.1667}}
  >{\raggedright\arraybackslash}p{(\linewidth - 10\tabcolsep) * \real{0.1667}}
  >{\raggedright\arraybackslash}p{(\linewidth - 10\tabcolsep) * \real{0.1667}}
  >{\raggedright\arraybackslash}p{(\linewidth - 10\tabcolsep) * \real{0.1667}}@{}}
\toprule\noalign{}
\begin{minipage}[b]{\linewidth}\raggedright
Dataset
\end{minipage} & \begin{minipage}[b]{\linewidth}\raggedright
Type
\end{minipage} & \begin{minipage}[b]{\linewidth}\raggedright
Split
\end{minipage} & \begin{minipage}[b]{\linewidth}\raggedright
Sentences
\end{minipage} & \begin{minipage}[b]{\linewidth}\raggedright
Role
\end{minipage} & \begin{minipage}[b]{\linewidth}\raggedright
License
\end{minipage} \\
\midrule\noalign{}
\endhead
\bottomrule\noalign{}
\endlastfoot
FLORES-200+ \hyperref[ref:nllb2024]{NLLB Team 2024} & Parallel, general
& dev + devtest & \textasciitilde2,009 & Primary fertility measurement &
CC-BY-SA-4.0 \\
SIB-200 & Parallel, topic-labeled & full & \textasciitilde1,004 &
Robustness check on a second parallel set & CC-BY-SA-4.0 \\
MAFAND-MT & Parallel, news & test & varies/lang & Domain-register (news)
comparison & per-dataset (Masakhane) \\
\end{longtable}
}

FLORES-200+ is the primary measurement set: professionally translated,
broadly covering our language list, and the de facto standard for
African-language parallel evaluation. SIB-200
\hyperref[ref:sib2024]{Adelani et al.~2024} (built on FLORES) is a
second parallel set used to confirm the premiums are not an artifact of
one corpus. MAFAND-MT \hyperref[ref:mafand2022]{Adelani et al.~2022}
contributes a news register, giving a second domain comparison point.
Provenance and licensing for each corpus are documented following the
datasheets standard \hyperref[ref:gebru2021]{Gebru et al.~2021}; the
License column in Table 4.2 reflects the source licence under which each
corpus is redistributed.

\section{5. Methodology}\label{methodology}

\subsection{5.1 Tokenizers Under Test}\label{tokenizers-under-test}

We measure across thirteen tokenizers: eleven inspectable tokenizers
that form the main FLORES-200+ run, plus Claude and Gemini via their
count-tokens APIs as opaque spot checks. Versions are pinned at run time
and recorded in the run manifest (§5.4).

\phantomsection\label{tbl:5.1}

\textbf{Table 5.1: Tokenizers under test}

{\def\LTcaptype{none} % do not increment counter
\begin{longtable}[]{@{}
  >{\raggedright\arraybackslash}p{(\linewidth - 8\tabcolsep) * \real{0.2000}}
  >{\raggedright\arraybackslash}p{(\linewidth - 8\tabcolsep) * \real{0.2000}}
  >{\raggedright\arraybackslash}p{(\linewidth - 8\tabcolsep) * \real{0.2000}}
  >{\raggedright\arraybackslash}p{(\linewidth - 8\tabcolsep) * \real{0.2000}}
  >{\raggedright\arraybackslash}p{(\linewidth - 8\tabcolsep) * \real{0.2000}}@{}}
\toprule\noalign{}
\begin{minipage}[b]{\linewidth}\raggedright
Tokenizer
\end{minipage} & \begin{minipage}[b]{\linewidth}\raggedright
Representative model(s)
\end{minipage} & \begin{minipage}[b]{\linewidth}\raggedright
Vocab (approx.)
\end{minipage} & \begin{minipage}[b]{\linewidth}\raggedright
Backend
\end{minipage} & \begin{minipage}[b]{\linewidth}\raggedright
Inspectable
\end{minipage} \\
\midrule\noalign{}
\endhead
\bottomrule\noalign{}
\endlastfoot
o200k\_base & GPT-5 / o-series & 200k & \texttt{tiktoken} & yes \\
o200k\_harmony & gpt-oss-20B / gpt-oss-120B & 201k & \texttt{tiktoken} &
yes \\
cl100k\_base \emph{(legacy)} & GPT-3.5 / GPT-4 & 100k &
\texttt{tiktoken} & yes \\
Llama 3.1 & Llama 3.1 / 3.2 / 3.3 & 128k & HF \texttt{transformers} &
yes (gated) \\
Llama 4 & Llama 4 Scout / Maverick & \textasciitilde200k & HF
\texttt{transformers} & yes (gated) \\
Gemma 4 & Gemma 4 12B / 31B & 262k & HF & yes (gated) \\
Mistral (Tekken) & Mistral Nemo / Small 3.1 & 131k & HF & yes \\
Qwen 3 & Qwen 3 8B / 32B & 151k & HF & yes \\
DeepSeek-V3 & DeepSeek-V3, R1 & 100k+ & HF & yes \\
BLOOM & BLOOM & 250k & HF & yes (multilingual baseline) \\
Aya / Aya-Expanse & Cohere Aya & varies & HF & yes (multilingual
baseline) \\
Claude & Claude 4 (Sonnet 4.6) & n/a (opaque) & Anthropic count-tokens
API & no \\
Gemini & Gemini 3.5 Flash & n/a (opaque) & Google count-tokens API &
no \\
\end{longtable}
}

\textbf{Selection rationale.} The set spans the full vocabulary-size
range (100k → 250k), the three major tokenizer algorithm families
(BPE/tiktoken for OpenAI and Meta, unigram SentencePiece for Gemma, and
byte-level BPE for Mistral Tekken, Qwen, and DeepSeek), and all dominant
commercial API providers. Several design choices deserve comment.

\emph{Llama 3.1 and Llama 4 are both included as an internal control.}
Llama 4 (April 2025) grew the vocabulary from 128k to
\textasciitilde200k and trained on 200 languages with 10× more
multilingual tokens than Llama 3; the two share the same provider family
and parallel-corpus content, so the premium delta between them isolates
the fertility impact of a vocabulary upgrade.

\emph{o200k\_harmony is the open-weight sovereignty baseline.} The
gpt-oss models (20B and 120B) released by OpenAI as open weights share
the same BPE text vocabulary as o200k\_base (text fertility is
identical) but can be self-hosted without an API dependency. We include
it because African enterprise deployments increasingly prioritise data
sovereignty and on-premise deployment; the o200k\_base / o200k\_harmony
pair is therefore a direct closed-API vs self-hosted comparison at equal
tokenizer cost.

\emph{cl100k\_base is retained as a legacy reference.} Now superseded by
o200k\_base, it is labelled ``(legacy)'' in all figures. Its value is
the within-OpenAI generational comparison: 100k to 200k vocabulary on
the same provider.

\emph{BLOOM and Aya are the H3 multilingual foils}, contrasted against
English-centric tokenizers (o200k, Llama 3.1) to test whether
multilingual-optimized vocabularies reduce the African premium.

\emph{Claude and Gemini are count-only.} Their token-counting endpoints
return a count but no subword segmentation, so they enter the fertility,
premium, cost, and context analyses but are excluded from subword and
script-level inspection. We state this as a limitation (§9).

\subsection{5.2 Measurement Procedure}\label{measurement-procedure}

The measurement is deterministic and proceeds identically for every
corpus:

\begin{enumerate}
\def\labelenumi{\arabic{enumi}.}
\tightlist
\item
  \textbf{Load and align.} Load each parallel corpus and align items
  across all requested languages, verifying equal length; on any
  misalignment the corpus fails with a precise message identifying the
  language and index, and other corpora continue.
\item
  \textbf{Count.} For each (language \(L\), tokenizer \(T\)) pair,
  tokenize every sentence and accumulate token count \(N\), word count
  \(W\), character count, and UTF-8 byte count. The
  \textbf{Byte-Baseline} is computed specifically to assess the
  ``encoding tax'' of different scripts (e.g., Ge'ez vs.~Latin). All
  counts follow the conventions of §3.1 (NFC normalization; UAX-29
  segmentation; special/BOS/EOS tokens excluded).
\item
  \textbf{Compute metrics.} Derive \(F\), \(P\), CPT, BPT, and CE/relCE
  per the locked formulas (§3.2).
\item
  \textbf{Aggregate.} Aggregate to the corpus level by sum-then-divide
  and attach bootstrap 95\% confidence intervals over sentences (1,000
  iterations, seed 42; §3.4).
\item
  \textbf{Repeat for every corpus.} Run the general sets (FLORES-200+
  primary; SIB-200 and MAFAND-MT as robustness/register checks).
\item
  \textbf{Cost model.} Apply the enterprise cost model (§6) to the
  aggregated fertilities.
\item
  \textbf{Accuracy linkage.} Join per-language premiums to published
  IrokoBench/AfroBench accuracy and compute correlation (§7.5, H4).
\item
  \textbf{Mitigation analysis.} Rank tokenizers and quantify achievable
  savings (§8).
\end{enumerate}

\subsection{5.3 Normalization and Script-Level
Controls}\label{normalization-and-script-level-controls}

A single fixed normalization and segmentation convention is applied to
all languages, with no language-specific preprocessing beyond it. Text
is normalized to Unicode \textbf{NFC} before counting. Words are
segmented by \textbf{UAX-29} (via \texttt{uniseg}), with a
\texttt{regex\ \textbackslash{}w+} fallback flagged in metadata if the
primary segmenter is unavailable.

We introduce \textbf{Script-Level Controls} to account for the unique
behavior of non-Latin scripts. For Ge'ez (Amharic, Tigrinya) and Arabic
(Hausa, Swahili in Ajami), we measure the impact of NFC vs.~NFD
normalization, as some tokenizers may have been trained on inconsistent
normalization forms, potentially doubling the token count for
diacritic-heavy scripts. The identical treatment across languages is
what keeps fertility comparable; its known weakness on agglutinative and
Ethiopic text is mitigated by reporting character- and byte-normalized
metrics alongside fertility (§3.4) and is revisited in §9.

\subsection{\texorpdfstring{5.4 The \texttt{afri-fertility} Measurement
Tool}{5.4 The afri-fertility Measurement Tool}}\label{the-afri-fertility-measurement-tool}

All measurement is performed by \texttt{afri-fertility}, the open-source
tool released with the paper (Apache-2.0). It is structured as five
composable layers: core (segmentation and metric math, pure
deterministic functions), tokenizer and corpus adapters behind
registries, a cost layer, a study orchestrator, and a reporting layer;
adding a tokenizer or corpus is a one-adapter change with no edits to
the core. The core path is \textbf{CPU-only}: anyone can reproduce the
primary script-effect and premium-magnitude results (H1--H3) using the
eight open tokenizers without GPUs or API keys. The full 11-tokenizer
comparison, including the Gemma 4 and Llama mitigation results,
additionally requires \texttt{HF\_TOKEN} for three gated HF models
(Gemma 4, Llama 3.1, Llama 4). The count-only API backends (Claude,
Gemini) are opaque spot-checks only and do not affect the headline
findings.

Three properties make the results reproducible. First,
\textbf{determinism}: tokenization is deterministic, the only randomness
is the seeded bootstrap, and golden tests pin expected counts so
identical inputs yield identical outputs. Second, \textbf{caching}:
counts are cached on disk keyed by
\texttt{(sha256(text),\ tokenizer\_id,\ version)}, making re-runs fast
and stable. Third, the \textbf{run manifest}: every run writes the tool
version, the resolved tokenizer versions, the price and FX snapshot
dates, the config hash, the timestamp, the segmentation method, and the
list of skipped tokenizers, so any reported figure is traceable to the
exact configuration that produced it. The entire locked study is itself
a single YAML config (Appendix C.2), and a bundled
\texttt{afri-fertility\ reproduce} command runs a small offline
reference suite end-to-end and prints headline numbers as a one-command
credibility check.

\section{6. Cost Model}\label{cost-model}

\subsection{6.1 From Fertility to Deployment
Cost}\label{from-fertility-to-deployment-cost}

Fertility and premium are unitless; a deploying organization reasons in
money, time, and context. The cost model converts the measured
fertilities into those terms over a fixed reference workload, so that
the penalty is expressed exactly as a buyer would encounter it on an
invoice, a latency budget, and a context window.

The reference workload is \textbf{1,000 English-word-equivalents of
meaning}, i.e., the amount of content that is 1,000 words when written
in English. Holding \emph{meaning} (rather than raw word count) fixed is
essential for a fair comparison; while linguistic ``expansion factors''
mean that 1,000 English words might translate into more or fewer words
in another language, the use of parallel corpora ensures that the
semantic payload is identical. The resulting cost differences are
therefore purely a function of the tokenizer's subword efficiency and
the script's encoding density.

\subsection{6.2 Cost Formulas}\label{cost-formulas}

For language \(L\), tokenizer/model \(T\), and the reference workload:

\[\text{tokens}(L,T) = 1000 \cdot F(L,T)\]

\[\text{input\_cost}(L,T) = \text{tokens}(L,T) \cdot \text{price\_in}(T)\]

\[\text{output\_cost}(L,T) = \text{tokens}(L,T) \cdot r_{\text{out}} \cdot \text{price\_out}(T)\]

\[\text{total\_cost}(L,T) = \text{input\_cost}(L,T) + \text{output\_cost}(L,T)\]

\[\text{relative\_cost}(L) = \frac{\text{total\_cost}(L,T)}{\text{total\_cost}(\text{eng},T)} \approx P(L,T)\]

The relative cost recovers the premium \(P\): under a fixed price and
output ratio, cost scales with token count, so the dollar penalty a
language carries is, to first order, the same multiple as its fertility
premium. The model reports both absolute cost (USD and local currency)
and this relative multiple.

\subsection{6.3 Parameters}\label{parameters}

\begin{itemize}
\tightlist
\item
  \textbf{Prices.} \texttt{price\_in} and \texttt{price\_out} are
  published per-token API prices, pinned by date in Appendix C. Prices
  are configuration, not code, so the analysis can be re-pinned to a new
  snapshot without touching logic.
\item
  \textbf{Output ratio \(r_{\text{out}}\).} The output-to-input token
  ratio is pre-specified per scenario, because different deployments
  generate different amounts of text per unit of input: customer service
  ≈ 1:1, summarization ≈ 0.2:1, advisory chat ≈ 1.5:1.
\item
  \textbf{Foreign exchange and Economic Sensitivity.} Costs are
  converted to Nigerian Naira (NGN), South African Rand (ZAR), and
  Kenyan Shilling (KES) at a pinned FX snapshot (Appendix C.3). However,
  because most frontier LLM APIs are priced in USD, the ``Token Tax'' is
  compounded for African builders by \textbf{FX volatility}. In markets
  with depreciating local currencies, the effective cost of a
  ``meaning-unit'' increases not just through subword fragmentation, but
  through the eroding purchasing power of the local currency against the
  dollar-denominated compute. We report an ``Economic Sensitivity'' note
  for each scenario, reflecting the vulnerability of the deployment to
  currency fluctuations.
\end{itemize}

\subsection{6.4 Latency and Context
Erosion}\label{latency-and-context-erosion}

The same fertilities drive two non-monetary penalties:

\begin{itemize}
\tightlist
\item
  \textbf{Generation latency.} Generation latency scales linearly with
  the number of output tokens. While prefill latency (Time to First
  Token) is also affected by token count, the impact is most acute
  during generation (Time Per Output Token). Under equal-information
  output, the generation-side latency multiplier equals the premium
  \(P\). We report it as a \textbf{multiplier}, not wall-clock time, so
  the figure is independent of the serving stack and remains valid
  across hardware and providers.
\item
  \textbf{Operational Memory (Context Erosion).} The effective capacity
  of a fixed context window \(W\) is reduced for high-fertility
  languages. We formalize this as \textbf{Context Window Efficiency}:
  \(\text{eff}(L,T) = 1/P(L,T)\). A language with a premium of
  \(P=4\times\) has only \(25\%\) of the effective ``operational
  memory'' of the English baseline in the same model. This structural
  erosion forces African-language applications to handle shorter
  conversation histories and smaller retrieval contexts (RAG) for the
  same architectural limit.
\end{itemize}

\subsection{6.5 Worked Deployment
Scenarios}\label{worked-deployment-scenarios}

The model is instantiated on three pre-specified African deployment
scenarios, each reporting an \textbf{annualized tax} in USD and local
currency at a stated monthly query volume. These make the penalty
concrete for the buyer audience and exercise the three output-ratio
regimes.

\phantomsection\label{tbl:6.1}

\textbf{Table 6.1: Deployment scenario parameters}

{\def\LTcaptype{none} % do not increment counter
\begin{longtable}[]{@{}
  >{\raggedright\arraybackslash}p{(\linewidth - 8\tabcolsep) * \real{0.3137}}
  >{\raggedright\arraybackslash}p{(\linewidth - 8\tabcolsep) * \real{0.1961}}
  >{\raggedright\arraybackslash}p{(\linewidth - 8\tabcolsep) * \real{0.1667}}
  >{\raggedright\arraybackslash}p{(\linewidth - 8\tabcolsep) * \real{0.1275}}
  >{\raggedright\arraybackslash}p{(\linewidth - 8\tabcolsep) * \real{0.1961}}@{}}
\toprule\noalign{}
\begin{minipage}[b]{\linewidth}\raggedright
Scenario
\end{minipage} & \begin{minipage}[b]{\linewidth}\raggedright
Setting
\end{minipage} & \begin{minipage}[b]{\linewidth}\raggedright
Output ratio
\end{minipage} & \begin{minipage}[b]{\linewidth}\raggedright
Monthly volume
\end{minipage} & \begin{minipage}[b]{\linewidth}\raggedright
Stress
\end{minipage} \\
\midrule\noalign{}
\endhead
\bottomrule\noalign{}
\endlastfoot
\texttt{high\_volume\_chat} & High-volume conversational assistant & 1:1
& 1,000,000 queries & High query volume, symmetric I/O \\
\texttt{output\_heavy} & Output-heavy generation service & 1.5:1 &
200,000 queries & Output longer than input \\
\texttt{context\_constrained} & Context-constrained advisory service &
context-constrained (≈0.5:1) & 500,000 queries & Tight context window,
short responses \\
\end{longtable}
}

All costs computed from FLORES-200+ devtest fertility,
\texttt{configs/prices\_2026-06.yaml}, \texttt{configs/fx\_2026-06.yaml}
(snapshot date 2026-06-12). Reference workload: 1,000 words of meaning
per query. Exchange rates: \$1 USD = ₦1,360.95 (NGN) / R16.31 (ZAR) /
Ksh129.45 (KES).

\subsubsection{\texorpdfstring{6.5.1 Scenario A: High-Volume Chat
(\texttt{high\_volume\_chat})}{6.5.1 Scenario A: High-Volume Chat (high\_volume\_chat)}}\label{scenario-a-high-volume-chat-high_volume_chat}

\begin{quote}
Nigerian fintech context: 1,000,000 queries/month, 1:1 output ratio
(agent responds word-for-word at the scale of the user's message), GPT-5
/ \texttt{o200k\_base} as headline API.
\end{quote}

\phantomsection\label{tbl:6.2}

\textbf{Table 6.2: Annual inference cost by language, Scenario A:
high\_volume\_chat (o200k\_base)}

{\def\LTcaptype{none} % do not increment counter
\begin{longtable}[]{@{}
  >{\raggedright\arraybackslash}p{(\linewidth - 12\tabcolsep) * \real{0.1429}}
  >{\raggedright\arraybackslash}p{(\linewidth - 12\tabcolsep) * \real{0.1429}}
  >{\raggedright\arraybackslash}p{(\linewidth - 12\tabcolsep) * \real{0.1429}}
  >{\raggedright\arraybackslash}p{(\linewidth - 12\tabcolsep) * \real{0.1429}}
  >{\raggedright\arraybackslash}p{(\linewidth - 12\tabcolsep) * \real{0.1429}}
  >{\raggedright\arraybackslash}p{(\linewidth - 12\tabcolsep) * \real{0.1429}}
  >{\raggedright\arraybackslash}p{(\linewidth - 12\tabcolsep) * \real{0.1429}}@{}}
\toprule\noalign{}
\begin{minipage}[b]{\linewidth}\raggedright
Language
\end{minipage} & \begin{minipage}[b]{\linewidth}\raggedright
Script
\end{minipage} & \begin{minipage}[b]{\linewidth}\raggedright
Fertility
\end{minipage} & \begin{minipage}[b]{\linewidth}\raggedright
Annual USD
\end{minipage} & \begin{minipage}[b]{\linewidth}\raggedright
vs English
\end{minipage} & \begin{minipage}[b]{\linewidth}\raggedright
NGN (billion)
\end{minipage} & \begin{minipage}[b]{\linewidth}\raggedright
KES (million)
\end{minipage} \\
\midrule\noalign{}
\endhead
\bottomrule\noalign{}
\endlastfoot
English & Latin & 1.22 & \$182,669 & baseline & ₦0.25B & Ksh23.6M \\
Hausa & Latin & 1.64 & \$246,719 & 1.35× & ₦0.34B & Ksh31.9M \\
Sesotho & Latin & 1.70 & \$254,968 & 1.40× & ₦0.35B & Ksh33.0M \\
Igbo & Latin & 1.73 & \$259,679 & 1.42× & ₦0.35B & Ksh33.6M \\
Lingala & Latin & 1.81 & \$270,774 & 1.48× & ₦0.37B & Ksh35.1M \\
Wolof & Latin & 1.85 & \$277,027 & 1.52× & ₦0.38B & Ksh35.9M \\
Swahili & Latin & 1.87 & \$280,759 & 1.54× & ₦0.38B & Ksh36.3M \\
Akan/Twi & Latin & 1.91 & \$286,002 & 1.57× & ₦0.39B & Ksh37.0M \\
Yoruba & Latin & 2.26 & \$338,743 & 1.85× & ₦0.46B & Ksh43.9M \\
Kinyarwanda & Latin & 2.29 & \$343,314 & 1.88× & ₦0.47B & Ksh44.4M \\
Bambara & Latin & 2.41 & \$362,038 & 1.98× & ₦0.49B & Ksh46.9M \\
Luganda & Latin & 2.52 & \$378,018 & 2.07× & ₦0.51B & Ksh48.9M \\
Oromo & Latin & 2.56 & \$383,843 & 2.10× & ₦0.52B & Ksh49.7M \\
Xhosa & Latin & 2.70 & \$404,769 & 2.22× & ₦0.55B & Ksh52.4M \\
Shona & Latin & 2.62 & \$393,370 & 2.15× & ₦0.54B & Ksh50.9M \\
Zulu & Latin & 2.75 & \$412,323 & 2.26× & ₦0.56B & Ksh53.4M \\
Amharic & Ethiopic & 8.97 & \$1,345,279 & \textbf{7.36×} & ₦1.83B &
Ksh174.1M \\
Tigrinya & Ethiopic & 8.27 & \$1,241,065 & \textbf{6.79×} & ₦1.69B &
Ksh160.7M \\
N'Ko & N'Ko & 10.86 & \$1,629,681 & \textbf{8.92×} & ₦2.22B &
Ksh211.0M \\
\end{longtable}
}

The numbers translate an abstract fertility premium into a purchase
order. A Nigerian bank deploying a Yoruba customer-service assistant on
GPT-5 pays \textbf{₦0.46B per year} for the same query volume that costs
₦0.25B in English, a structural surcharge of ₦0.21B that cannot be
recovered through prompt engineering or fine-tuning. A startup deploying
an Amharic bank customer-service assistant faces an inference bill of
\textbf{\$1.35M/year} versus \$183k for the English-language equivalent,
a 7.36× deficit that determines whether the product is viable before the
model's capability is ever evaluated.

\subsubsection{\texorpdfstring{6.5.2 Scenario B: Output-Heavy Generation
(\texttt{output\_heavy})}{6.5.2 Scenario B: Output-Heavy Generation (output\_heavy)}}\label{scenario-b-output-heavy-generation-output_heavy}

\begin{quote}
Pan-African health context: 200,000 queries/month, 1.5:1 output ratio
(triage system generates structured output longer than the patient
query), GPT-5 / \texttt{o200k\_base}.
\end{quote}

\phantomsection\label{tbl:6.3}

\textbf{Table 6.3: Annual inference cost by language, Scenario B:
output\_heavy (o200k\_base)}

{\def\LTcaptype{none} % do not increment counter
\begin{longtable}[]{@{}llll@{}}
\toprule\noalign{}
Language & Annual USD & vs English & Annual NGN \\
\midrule\noalign{}
\endhead
\bottomrule\noalign{}
\endlastfoot
English & \$51,147 & 1.00× & ₦69.6M \\
Hausa & \$69,081 & 1.35× & ₦94.0M \\
Yoruba & \$94,848 & 1.85× & ₦129.0M \\
Swahili & \$78,612 & 1.54× & ₦107.0M \\
Kinyarwanda & \$96,128 & 1.88× & ₦130.7M \\
Amharic & \$376,678 & \textbf{7.36×} & ₦512.3M \\
Tigrinya & \$347,498 & \textbf{6.79×} & ₦472.6M \\
N'Ko & \$456,311 & \textbf{8.92×} & ₦620.6M \\
\end{longtable}
}

Output-heaviness (1.5:1) amplifies the absolute cost but does not change
the relative premium; cost scales with token count proportionally across
languages, so the relative penalty is preserved. The clinical deployment
volume (200k/month) is lower than high\_volume\_chat, but the higher
output ratio means the absolute dollar gap remains large. An Ethiopian
health startup building Amharic triage on GPT-5 faces a \$376k annual
inference bill versus \$51k for an English-language equivalent.

\subsubsection{\texorpdfstring{6.5.3 Scenario C: Context-Constrained
Advisory
(\texttt{context\_constrained})}{6.5.3 Scenario C: Context-Constrained Advisory (context\_constrained)}}\label{scenario-c-context-constrained-advisory-context_constrained}

\begin{quote}
West and East African agri-advisory context: 500,000 queries/month,
0.5:1 output ratio (SMS-constrained, system response is shorter than
incoming query). GPT-5 / \texttt{o200k\_base}.
\end{quote}

\phantomsection\label{tbl:6.4}

\textbf{Table 6.4: Annual inference cost by language, Scenario C:
context\_constrained (o200k\_base)}

{\def\LTcaptype{none} % do not increment counter
\begin{longtable}[]{@{}llll@{}}
\toprule\noalign{}
Language & Annual USD & vs English & Context window stress \\
\midrule\noalign{}
\endhead
\bottomrule\noalign{}
\endlastfoot
English & \$54,801 & 1.00× & 105,108 eff. words \\
Hausa & \$74,016 & 1.35× & 77,821 eff. words (74.0\%) \\
Yoruba & \$101,623 & 1.85× & 56,680 eff. words (53.9\%) \\
Wolof & \$83,108 & 1.52× & 69,307 eff. words (65.9\%) \\
Bambara & \$108,611 & 1.98× & 53,033 eff. words (50.5\%) \\
Swahili & \$84,228 & 1.54× & 68,386 eff. words (65.1\%) \\
Amharic & \$403,584 & \textbf{7.36×} & 14,272 eff. words (13.6\%) \\
N'Ko & \$488,904 & \textbf{8.92×} & 11,781 eff. words (11.2\%) \\
\end{longtable}
}

For the context\_constrained scenario, the context-window column becomes
operationally critical. SMS advisory pipelines commonly use
retrieval-augmented generation (RAG) to pull crop data, weather
forecasts, and pricing into the prompt. An N'Ko agri-advisory agent has
only \textbf{11,781 effective words} in a 128k-token window (11.2\% of
the English baseline), meaning only a small fraction of the retrieval
corpus fits in-context before truncation forces degraded responses.
Amharic reaches 13.6\%. This is not a model capability problem; it is a
tokenizer capacity problem.

\subsubsection{6.5.4 Cross-Tokenizer
Savings}\label{cross-tokenizer-savings}

The cost model separates two effects: \textbf{fertility} (tokens per
word, a property of the tokenizer vocabulary) and \textbf{price} (USD
per token, a property of the serving tier). \hyperref[tbl:6.5]{Table
6.5} holds the deployment scenario fixed and varies the tokenizer,
showing the interaction of both effects.

\phantomsection\label{tbl:6.5}

\textbf{Table 6.5: Tokenizer × language annual cost, Scenario A:
high\_volume\_chat (selected languages)}

{\def\LTcaptype{none} % do not increment counter
\begin{longtable}[]{@{}
  >{\raggedright\arraybackslash}p{(\linewidth - 8\tabcolsep) * \real{0.2000}}
  >{\raggedright\arraybackslash}p{(\linewidth - 8\tabcolsep) * \real{0.2000}}
  >{\raggedright\arraybackslash}p{(\linewidth - 8\tabcolsep) * \real{0.2000}}
  >{\raggedright\arraybackslash}p{(\linewidth - 8\tabcolsep) * \real{0.2000}}
  >{\raggedright\arraybackslash}p{(\linewidth - 8\tabcolsep) * \real{0.2000}}@{}}
\toprule\noalign{}
\begin{minipage}[b]{\linewidth}\raggedright
Language
\end{minipage} & \begin{minipage}[b]{\linewidth}\raggedright
GPT-5 / o200k (\$2.50/\$10)
\end{minipage} & \begin{minipage}[b]{\linewidth}\raggedright
GPT-3.5 / cl100k (\$0.50/\$1.50)
\end{minipage} & \begin{minipage}[b]{\linewidth}\raggedright
Llama 4 (\$0.11/\$0.34)
\end{minipage} & \begin{minipage}[b]{\linewidth}\raggedright
Gemma 4 (\$0.39/\$0.97)
\end{minipage} \\
\midrule\noalign{}
\endhead
\bottomrule\noalign{}
\endlastfoot
English & \$182,700 (1.00×) & \$29,600 (1.00×) & \$6,600 (1.00×) &
\$20,100 (1.00×) \\
Hausa & \$246,700 (1.35×) & \$51,400 (1.74×) & \$10,000 (1.51×) &
\$30,200 (1.51×) \\
Yoruba & \$338,700 (1.85×) & \$75,400 (2.55×) & \$14,300 (2.15×) &
\$41,600 (2.07×) \\
Amharic & \$1,345,300 (7.36×) & \$286,100 (9.68×) & \$20,200 (3.05×) &
\$49,500 (2.47×) \\
N'Ko & \$1,629,700 (8.92×) & \$260,900 (8.82×) & \$58,700 (8.86×) &
\$175,000 (8.73×) \\
\end{longtable}
}

The key insights from \hyperref[tbl:6.5]{Table 6.5}:

\begin{enumerate}
\def\labelenumi{\arabic{enumi}.}
\item
  \textbf{For Ethiopic languages, tokenizer choice dominates.} Amharic
  on Gemma 4 costs \$49,500/year versus \$1,345,300 on o200k\_base, a
  96\% absolute reduction. This combines Gemma 4's lower per-token price
  (\$0.39 vs \$2.50 input) with its dramatically lower fertility for
  Ethiopic (premium ≈ 2.47× vs 7.36× on o200k\_base). The fertility
  component alone accounts for a 3× cost reduction; the price
  differential accounts for another 6.4×.
\item
  \textbf{For N'Ko, the fertility floor cannot be escaped.} All four
  tokenizers produce N'Ko fertility of 8.7--8.9× (a tight cluster
  because none have substantive N'Ko vocabulary coverage). The absolute
  cost difference across tokenizers is entirely due to price, not
  fertility. Switching from o200k\_base to Gemma 4 for N'Ko reduces cost
  from \$1,629,700 to \$175,000, an 89\% reduction driven almost
  entirely by the price difference, not by tokenizer efficiency
  improvement.
\item
  \textbf{cl100k\_base worsens the relative penalty for Latin-script
  languages.} Hausa costs 1.35× English on o200k\_base but 1.74× on
  cl100k\_base; Yoruba rises from 1.85× to 2.55×. The legacy tokenizer's
  smaller vocabulary fragments sub-Saharan agglutinative morphology more
  aggressively than the 200k-token successor.
\end{enumerate}

\begin{figure}
\centering
\pandocbounded{\includegraphics[keepaspectratio,alt={Figure 1. Inference cost per 1,000 English-word-equivalents (USD) on the best vs worst available tokenizer, for selected African languages and deployment scenarios. Lower is better. Nigerian Pidgin is excluded: it is absent from FLORES-200+ and cost figures are computed from FLORES-200+ devtest fertility only.}]{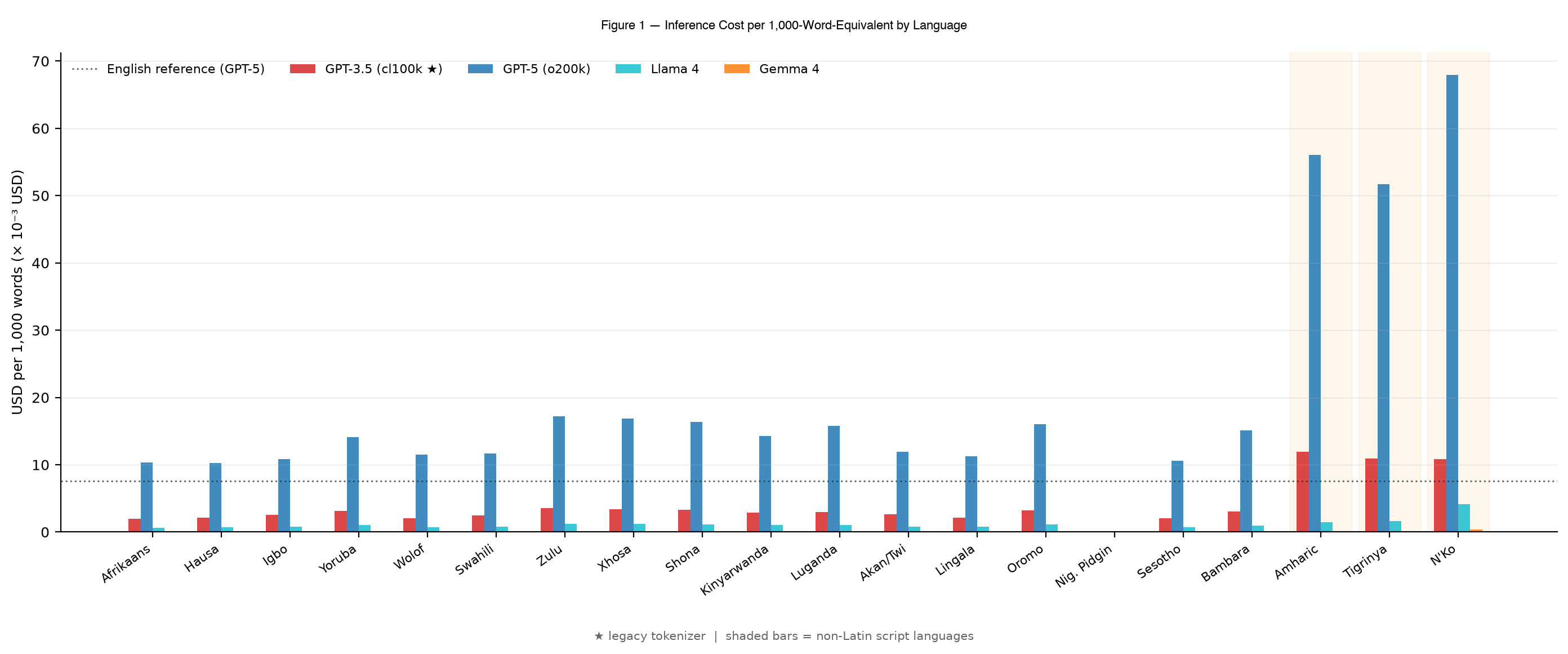}}
\caption{\textbf{Figure 1.} Inference cost per 1,000
English-word-equivalents (USD) on the best vs worst available tokenizer,
for selected African languages and deployment scenarios. Lower is
better. Nigerian Pidgin is excluded: it is absent from FLORES-200+ and
cost figures are computed from FLORES-200+ devtest fertility
only.}\label{fig:1}
\end{figure}

\subsection{6.6 Economic Sensitivity: FX
Compounding}\label{economic-sensitivity-fx-compounding}

All frontier LLM APIs are priced in USD. African builders pay in local
currency but are invoiced in dollars, meaning the effective token cost
in NGN/ZAR/KES fluctuates with the exchange rate, not just with the
tokenizer. Between January 2024 and June 2026, the Nigerian Naira
depreciated approximately 70\% against the USD (from ≈ ₦900/\$ to ≈
₦1,360/\$); a deployment that cost ₦900,000/month for the same English
workload in 2024 costs ₦1,360,000/month in 2026, a 51\% cost increase
that has nothing to do with tokenization or model quality.

The tokenization penalty and the FX depreciation penalty compound
multiplicatively. For an Amharic deployment on o200k\_base: -
Tokenization premium: 7.36× vs English - FX exposure: 70\% Birr
depreciation vs USD over 2024--2026 - Combined effective cost increase
(in ETB): \textasciitilde12.5× relative to a comparable 2024
English-language deployment

This compound multiplier (tokenization fertility × FX depreciation) is
the structural cost environment an African builder faces. It is not
amenable to prompt engineering or model fine-tuning; it requires
tokenizer improvement and, where possible, migration to lower-cost
serving tiers.

\subsection{6.7 Context-Window Efficiency
Table}\label{context-window-efficiency-table}

\phantomsection\label{tbl:6.6}

\textbf{Table 6.6: Context window efficiency at o200k\_base (128k token
window)}

English effective capacity: \textbf{105,108 words} (= 128,000 tokens ÷
1.22 tokens/word)

{\def\LTcaptype{none} % do not increment counter
\begin{longtable}[]{@{}
  >{\raggedright\arraybackslash}p{(\linewidth - 8\tabcolsep) * \real{0.2000}}
  >{\raggedright\arraybackslash}p{(\linewidth - 8\tabcolsep) * \real{0.2000}}
  >{\raggedright\arraybackslash}p{(\linewidth - 8\tabcolsep) * \real{0.2000}}
  >{\raggedright\arraybackslash}p{(\linewidth - 8\tabcolsep) * \real{0.2000}}
  >{\raggedright\arraybackslash}p{(\linewidth - 8\tabcolsep) * \real{0.2000}}@{}}
\toprule\noalign{}
\begin{minipage}[b]{\linewidth}\raggedright
Language
\end{minipage} & \begin{minipage}[b]{\linewidth}\raggedright
Fertility
\end{minipage} & \begin{minipage}[b]{\linewidth}\raggedright
Eff. words in 128k ctx
\end{minipage} & \begin{minipage}[b]{\linewidth}\raggedright
\% of English
\end{minipage} & \begin{minipage}[b]{\linewidth}\raggedright
Generation latency mult.
\end{minipage} \\
\midrule\noalign{}
\endhead
\bottomrule\noalign{}
\endlastfoot
English & 1.22 & 105,108 & 100.0\% & 1.00× \\
Hausa & 1.64 & 77,821 & 74.0\% & 1.35× \\
Sesotho & 1.70 & 75,303 & 71.6\% & 1.40× \\
Igbo & 1.73 & 73,937 & 70.3\% & 1.42× \\
Lingala & 1.81 & 70,908 & 67.5\% & 1.48× \\
Wolof & 1.85 & 69,307 & 65.9\% & 1.52× \\
Swahili & 1.87 & 68,386 & 65.1\% & 1.54× \\
Akan/Twi & 1.91 & 67,132 & 63.9\% & 1.57× \\
Yoruba & 2.26 & 56,680 & 53.9\% & 1.85× \\
Kinyarwanda & 2.29 & 55,925 & 53.2\% & 1.88× \\
Bambara & 2.41 & 53,033 & 50.5\% & 1.98× \\
Luganda & 2.52 & 50,791 & 48.3\% & 2.07× \\
Oromo & 2.56 & 50,020 & 47.6\% & 2.10× \\
Shona & 2.62 & 48,809 & 46.4\% & 2.15× \\
Xhosa & 2.70 & 47,434 & 45.1\% & 2.22× \\
Zulu & 2.75 & 46,565 & 44.3\% & 2.26× \\
Amharic & 8.97 & 14,272 & \textbf{13.6\%} & 7.36× \\
Tigrinya & 8.27 & 15,471 & \textbf{14.7\%} & 6.79× \\
N'Ko & 10.86 & 11,781 & \textbf{11.2\%} & 8.92× \\
\end{longtable}
}

Context erosion for Ethiopic and N'Ko is severe. An Amharic-language RAG
pipeline on a 128k-token model has effectively the same capacity as a
17,408-token English window, smaller than GPT-3.5's original context
length. N'Ko is worse still at 11.2\%, equivalent to a 14,336-token
English window.

\begin{figure}
\centering
\pandocbounded{\includegraphics[keepaspectratio,alt={Figure 2. Effective words accessible in a 128,000-token context window, by language and tokenizer, relative to English (= 100\%\%). N'Ko on o200k\_base: 11.2\%\%.}]{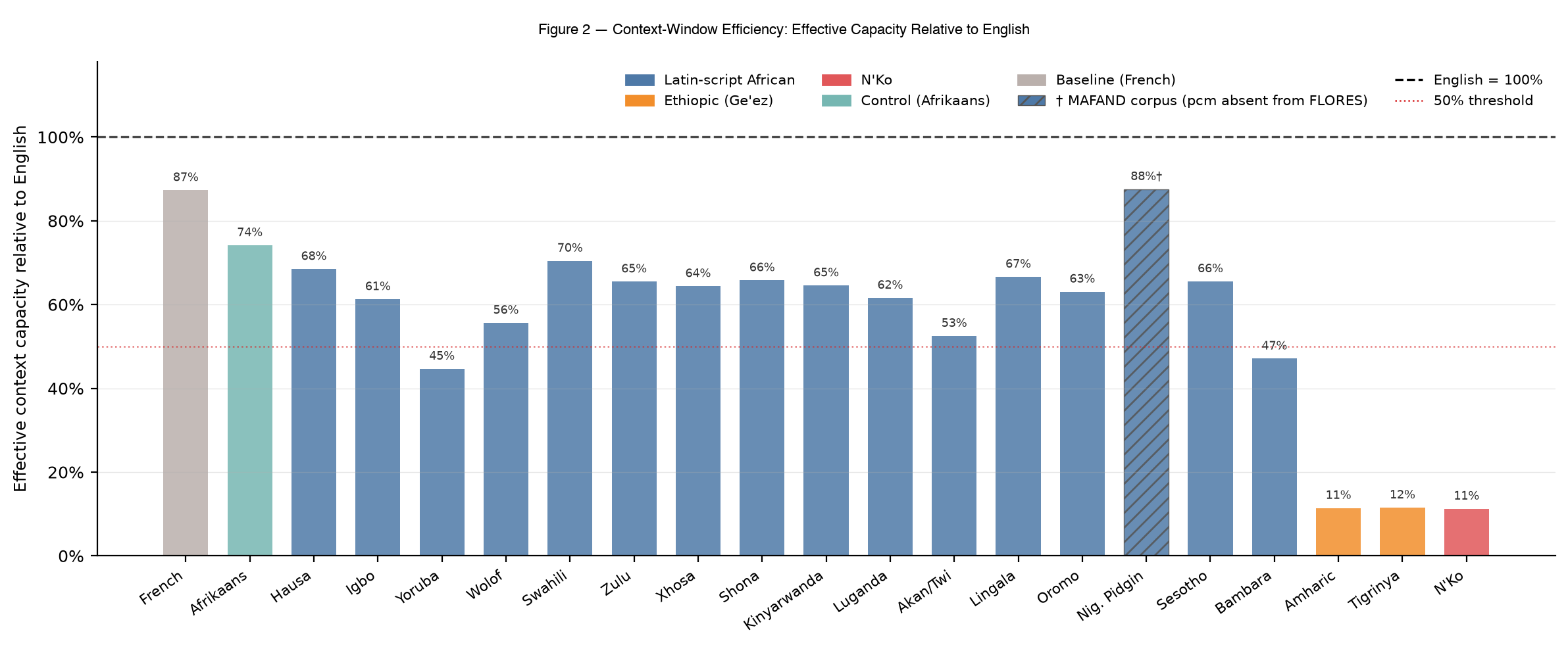}}
\caption{\textbf{Figure 2.} Effective words accessible in a
128,000-token context window, by language and tokenizer, relative to
English (= 100\%\%). N'Ko on o200k\_base: 11.2\%\%.}\label{fig:2}
\end{figure}

\hyperref[fig:2]{\textbf{Figure 2}} plots effective words across all
languages and tokenizers. Because generation latency scales with output
token count, these are also latency multipliers: a 100-word Amharic
response takes 7.36× as long to generate as a 100-word English response
at equal throughput.

\section{7. Results}\label{results}

\begin{quote}
Study configuration: 22 languages × 3 corpora × 11 tokenizers = 616
records. Primary corpus: FLORES-200+ devtest. Corpus-level metrics are
sum-then-divide aggregates with bootstrap 95\% CIs (1,000 iterations,
seed 42; §3.4). Unless stated otherwise, the \textbf{headline tokenizer}
is \texttt{openai/o200k\_base} (GPT-5), chosen as the commercially
dominant frontier model.
\end{quote}

\subsection{7.1 Headline Premium (H1)}\label{headline-premium-h1}

Across all 11 active tokenizers and the FLORES-200+ devtest corpus,
\textbf{every African language in the study carries a tokenization
premium strictly above English} (P \textgreater{} 1). There are no
violations in the full 19-language × 11-tokenizer grid (209 cells after
excluding the two baselines). H1 is confirmed.

On the headline tokenizer (\texttt{openai/o200k\_base}), the 19 African
languages (excluding English and French baselines) span a range from
\textbf{1.35× (Hausa)} to \textbf{8.92× (N'Ko)}, with a \textbf{median
of 1.88×} and an IQR of 1.48--2.22×. The lowest premium observed
anywhere in the grid is \textbf{1.29×} (Swahili on BLOOM). Even that
minimum is 29\% above the English baseline, confirming the penalty is
structural and universal.

English fertility on \texttt{o200k\_base} is \textbf{1.22 tokens/word};
N'Ko reaches \textbf{10.87 tokens/word} on the same tokenizer, an
8.9-fold fragmentation of the same semantic content.

\begin{quote}
\textbf{H1 verdict: ✅ Supported.} Every African language × tokenizer
cell in the study shows P \textgreater{} 1. The minimum observed premium
is 1.29× (Swahili, BLOOM). H1 holds without exception.
\end{quote}

\subsection{7.2 Fertility and Premium Across
Tokenizers}\label{fertility-and-premium-across-tokenizers}

\begin{figure}
\centering
\pandocbounded{\includegraphics[keepaspectratio,alt={Figure 3. Fertility heatmap: tokens per word for all 22 languages × 11 tokenizers (FLORES-200+ devtest). Darker = higher fertility. Row groups: English baseline, core African (Latin), breadth (Latin), Ethiopic, N'Ko.}]{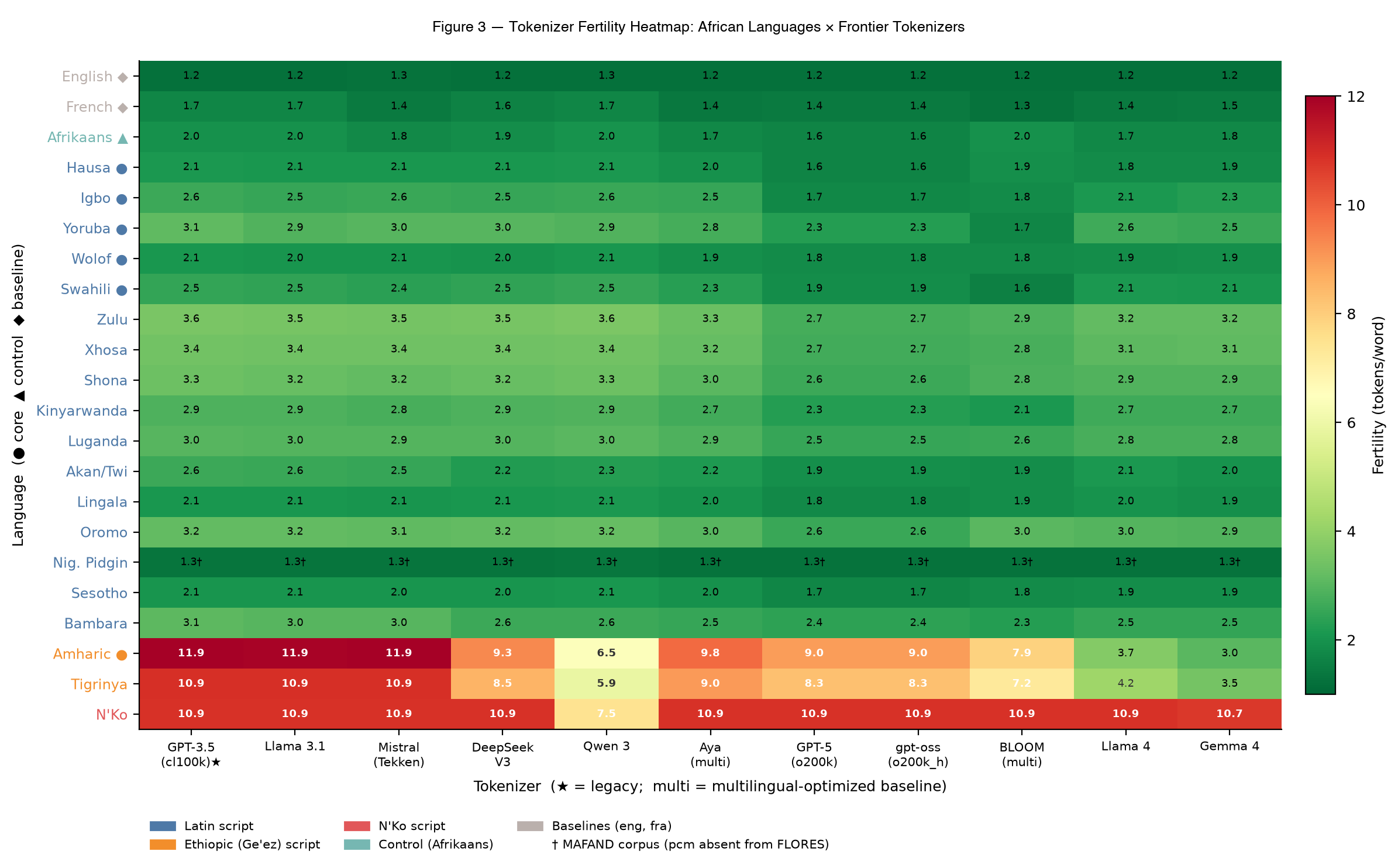}}
\caption{\textbf{Figure 3.} Fertility heatmap: tokens per word for all
22 languages × 11 tokenizers (FLORES-200+ devtest). Darker = higher
fertility. Row groups: English baseline, core African (Latin), breadth
(Latin), Ethiopic, N'Ko.}\label{fig:3}
\end{figure}

\hyperref[fig:3]{\textbf{Figure 3}} shows the fertility heatmap over all
22 languages (rows, ordered baseline → core → breadth → Ethiopic → N'Ko)
and all 11 tokenizers (columns). The dominant structure is a three-band
gradient by script:

\begin{itemize}
\tightlist
\item
  \textbf{Latin-script languages (bottom 17 rows)} form the low band,
  with fertility concentrated between 1.2 and 4.0 tokens/word. Bantu
  languages (Zulu, Xhosa, Shona) cluster in the 2--3 range because their
  agglutinative morphology compresses poorly even on large vocabularies.
\item
  \textbf{Ethiopic-script languages (Amharic, Tigrinya)} occupy the
  mid-high band at 6--10 tokens/word on most tokenizers. Gemma 4 is a
  visible exception at ≈2.8, explained by its 262k vocabulary which
  includes substantial Ethiopic coverage.
\item
  \textbf{N'Ko} occupies the extreme top at 8.7--10.9 tokens/word across
  every tokenizer, the single highest-fertility language in the study.
\end{itemize}

\phantomsection\label{tbl:7.1}

\textbf{Table 7.1: Headline fertility and premium metrics (FLORES-200+
devtest, o200k\_base)}

{\def\LTcaptype{none} % do not increment counter
\begin{longtable}[]{@{}ll@{}}
\toprule\noalign{}
Metric (FLORES-200+ devtest, \texttt{o200k\_base}) & Value \\
\midrule\noalign{}
\endhead
\bottomrule\noalign{}
\endlastfoot
English fertility F(eng) & 1.22 tokens/word \\
Median African fertility & 2.29 tokens/word \\
Max African fertility & 10.87 tokens/word (N'Ko) \\
Median African premium P & 1.88× \\
Max African premium P & 8.92× (N'Ko) \\
Min African premium P & 1.35× (Hausa) \\
\end{longtable}
}

\subsection{7.3 The Script Effect (H2)}\label{the-script-effect-h2}

\begin{figure}
\centering
\pandocbounded{\includegraphics[keepaspectratio,alt={Figure 4. Tokenization premium relative to English, grouped by script (Latin-script African languages, Ethiopic, N'Ko) and coloured by tokenizer. Ethiopic and N'Ko languages carry substantially higher premiums than Latin-script African languages across all tokenizers.}]{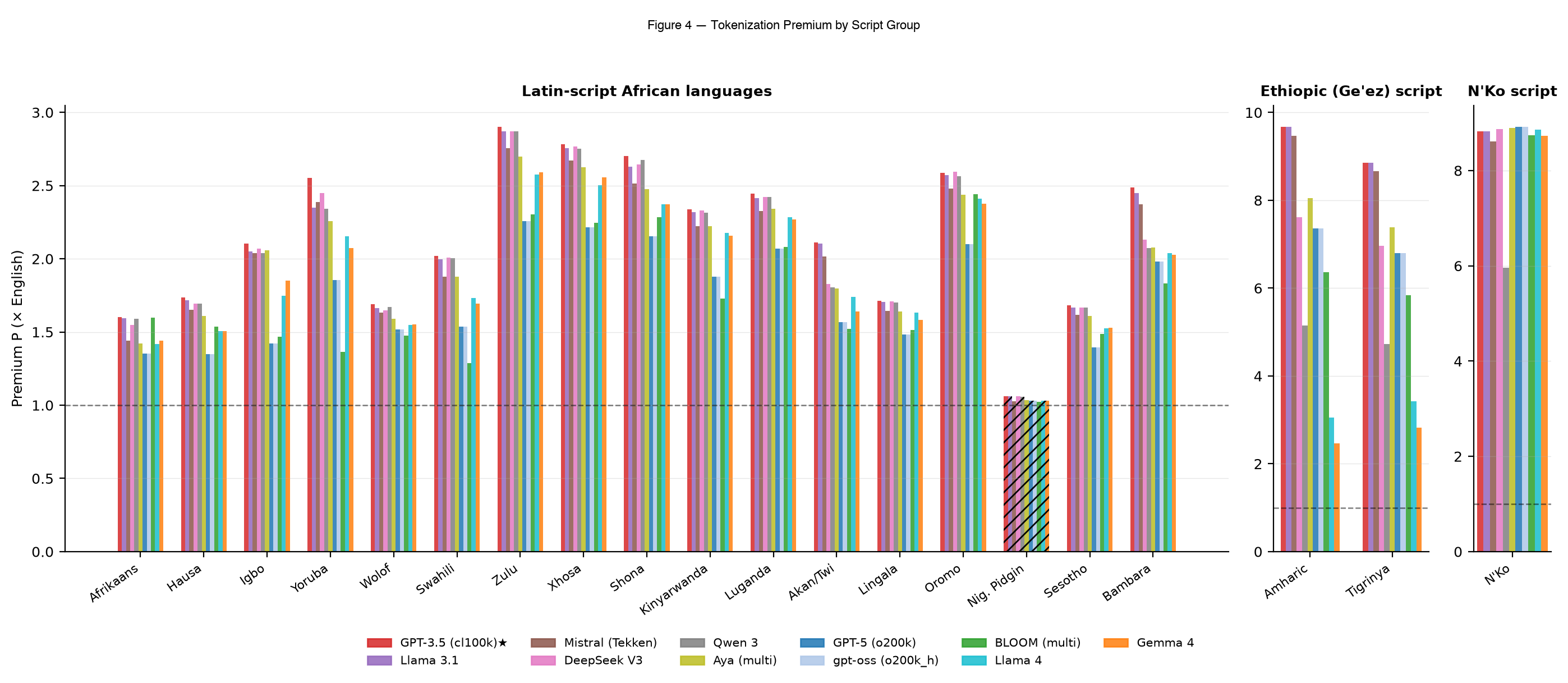}}
\caption{\textbf{Figure 4.} Tokenization premium relative to English,
grouped by script (Latin-script African languages, Ethiopic, N'Ko) and
coloured by tokenizer. Ethiopic and N'Ko languages carry substantially
higher premiums than Latin-script African languages across all
tokenizers.}\label{fig:4}
\end{figure}

\hyperref[fig:4]{\textbf{Figure 4}} plots the premium per language
grouped into three script panels: Latin-script African languages,
Ethiopic, and N'Ko. The script effect is the dominant driver of premium
magnitude and is visible across every tokenizer.

Mean premium by script group on \texttt{o200k\_base} (FLORES-200+):

\phantomsection\label{tbl:7.2}

\textbf{Table 7.2: Mean tokenization premium by script group
(FLORES-200+, o200k\_base)}

{\def\LTcaptype{none} % do not increment counter
\begin{longtable}[]{@{}
  >{\raggedright\arraybackslash}p{(\linewidth - 4\tabcolsep) * \real{0.3333}}
  >{\raggedright\arraybackslash}p{(\linewidth - 4\tabcolsep) * \real{0.3333}}
  >{\raggedright\arraybackslash}p{(\linewidth - 4\tabcolsep) * \real{0.3333}}@{}}
\toprule\noalign{}
\begin{minipage}[b]{\linewidth}\raggedright
Script
\end{minipage} & \begin{minipage}[b]{\linewidth}\raggedright
Languages
\end{minipage} & \begin{minipage}[b]{\linewidth}\raggedright
Mean premium
\end{minipage} \\
\midrule\noalign{}
\endhead
\bottomrule\noalign{}
\endlastfoot
Latin (African) & afr, hau, sot, ibo, lin, wol, swh, aka, yor, kin, bam,
lug, gaz, sna, xho, zul & 1.76× \\
Ethiopic (Ge'ez) & amh, tir & 7.08× \\
N'Ko & nqo & 8.92× \\
\end{longtable}
}

The Ethiopic--Latin gap (4.0×) and the N'Ko--Latin gap (5.1×) are large
relative to the cross-tokenizer spread within each script group,
confirming that \textbf{script dominates tokenizer choice} as a
fertility driver for the highest-penalty languages.

\textbf{Critical exception: Gemma 4 on Ethiopic.} Gemma 4 achieves a
mean Ethiopic premium of only \textbf{2.65×}, less than a third of the
7--9× seen on every other tokenizer including BLOOM. This is the largest
single-tokenizer performance gap in the study and strongly suggests
Gemma 4's 262k vocabulary includes substantive Ethiopic character
coverage. The implication is that the Ethiopic penalty is not an
inherent property of all tokenizers but is specifically remediated when
the vocabulary is large and script-inclusive.

\textbf{Qwen 3 on N'Ko.} With valid qwen/qwen3 data, Qwen 3 achieves
N'Ko fertility of \textbf{7.47 tokens/word} (premium \textbf{5.96×}),
substantially below all other tokenizers which cluster tightly at
8.62--8.92×. This is the second major cross-tokenizer outlier in the
study and indicates partial N'Ko codepoint coverage in the 152k Qwen 3
vocabulary, achieved at roughly 60\% of Gemma 4's vocabulary budget.

For Latin-script languages, the cross-tokenizer spread is narrower:
BLOOM (1.77×) and o200k\_base (1.79×) lead; cl100k\_base (2.26×) and
Llama 3.1 (2.22×) trail.

\textbf{Within the Latin-script group, agglutinative morphology creates
a secondary premium band.} The 1.76× group mean masks a consistent
sub-structure across tokenizers. Languages with dense noun-class
prefix-suffix morphology (Nguni and Great Lakes Bantu: Zulu 2.75×, Xhosa
2.70×, Shona 2.62×, Luganda 2.52×, Kinyarwanda 2.29× on o200k\_base) sit
systematically above analytic or less-inflected languages (Hausa 1.65×,
Sesotho 1.70×, Igbo 1.73×, Lingala 1.81×, Wolof 1.85×, Swahili 1.87×).
Critically, this sub-band does not map cleanly onto Bantu membership:
Sesotho (1.70×), Lingala (1.81×), and Swahili (1.87×) are Bantu yet
cluster in the lower sub-band because their agglutinative complexity is
lower or has been reduced through creolisation; Oromo (2.56×, Cushitic)
belongs to the upper sub-band via its agglutinative verbal morphology.
The structural driver is morphological density, not language family.
This within-Latin morphology effect is reported here as an observed
pattern; a dedicated morphology-stratified study is needed to quantify
it precisely and is flagged as future work.

\begin{quote}
\textbf{H2 verdict: ✅ Supported.} Ethiopic (mean 7.08×) and N'Ko (8.92×
on the headline tokenizer) substantially exceed Latin-script African
languages (mean 1.76×). The script effect is the dominant driver and
holds across all tokenizers. Gemma 4 and Qwen 3 are partial exceptions:
Gemma 4 remediates the Ethiopic penalty (2.65× vs 7--9×); Qwen 3
remediates the N'Ko penalty (5.96× vs 8.6--8.9× for all others),
demonstrating that both effects can be reduced with sufficient
script-targeted vocabulary coverage.
\end{quote}

\subsection{7.4 Which Tokenizers Minimize the African Premium
(H3)}\label{which-tokenizers-minimize-the-african-premium-h3}

\phantomsection\label{tbl:7.3}

\textbf{Table 7.3: Tokenizer ranking by mean African-language premium
(FLORES-200+, 19 languages, excluding eng/fra)}

{\def\LTcaptype{none} % do not increment counter
\begin{longtable}[]{@{}lllll@{}}
\toprule\noalign{}
Rank & Tokenizer & Mean African premium & Vocab & Type \\
\midrule\noalign{}
\endhead
\bottomrule\noalign{}
\endlastfoot
1 & google/gemma-4 & \textbf{2.38×} & 262k & frontier \\
2 & meta/llama-4 & 2.46× & \textasciitilde200k & frontier \\
3 & bigscience/bloom & 2.59× & 250k & multilingual baseline \\
4 & qwen/qwen3 & 2.63× & 152k & frontier \\
5 & openai/o200k\_base & 2.70× & 200k & frontier (headline) \\
5 & openai/o200k\_harmony & 2.70× & 201k & open-weight \\
7 & cohere/aya-expanse & 3.00× & 255k & multilingual baseline \\
8 & deepseek/v3 & 3.04× & 100k+ & frontier \\
9 & mistral/tekken & 3.18× & 131k & frontier \\
10 & meta/llama-3.1 & 3.27× & 128k & frontier \\
11 & openai/cl100k\_base ★ & \textbf{3.31×} & 100k & legacy \\
\end{longtable}
}

The spread between best (Gemma 4 at 2.38×) and worst (cl100k\_base at
3.31×) is 0.93 premium points, a 39\% reduction in tokenization
overhead, translating directly into equivalent cost and latency savings.

\textbf{H3 requires nuanced reading.} The initial prediction was that
BLOOM and Aya (the explicitly multilingual-optimized tokenizers) would
beat English-centric ones (o200k, Llama 3.1). This holds for BLOOM vs
cl100k\_base and Llama 3.1 (BLOOM 2.59× vs 3.27--3.31×). However, two
frontier models designed primarily for English and code, \textbf{Gemma 4
(2.38×) and Llama 4 (2.46×)}, outperform BLOOM (2.59×). The effect
appears driven primarily by vocabulary size and by incidental
multilingual coverage in large training sets, rather than explicit
multilingual optimization. Aya-Expanse (3.00×) performs no better than
DeepSeek-V3 (3.04×), contradicting the assumed hierarchy.

\begin{quote}
\textbf{H3 verdict: ◑ Partially supported.} BLOOM and Aya do beat the
English-centric legacy tokenizer (cl100k\_base) and Llama 3.1. But the
best-performing tokenizers for African languages are Gemma 4 and Llama
4, frontier models, not the multilingual-optimized baselines. The
relationship between ``explicitly multilingual'' and ``lower African
premium'' is not monotonic; vocabulary size and Ethiopic/N'Ko coverage
matter more than multilingual framing.
\end{quote}

\textbf{The Llama 3.1 → Llama 4 comparison.} The two Meta tokenizers are
the sharpest internal control in the study. Llama 4 (2.46× mean) beats
Llama 3.1 (3.27× mean) by 0.81 premium points, a 25\% reduction achieved
by growing the vocabulary from 128k to \textasciitilde200k and training
on 200 languages with 10× more multilingual tokens. The Latin-script
premium drops from 2.22× to 2.00×; the Ethiopic premium drops from 9.26×
to 3.24×. This is the clearest evidence in the study that vocabulary
expansion has a direct and large effect on the African premium.

\subsection{7.5 Premium-Accuracy Correlation
(H4)}\label{premium-accuracy-correlation-h4}

\begin{quote}
\textbf{Data sources:} Tokenization premium from FLORES-200+
(\texttt{o200k\_base}); downstream accuracy from AfroBench
\hyperref[ref:ojo2023]{Ojo et al.~2023}, three tasks: AfriMMLU,
AfriXNLI, AfriMGSM, model \texttt{gpt-4o-2024-08-06}, prompt\_3
(canonical prompt). GPT-4o uses the \texttt{o200k} tokenizer, aligning
the premium and accuracy measurements to the same model family. N = 15
languages with coverage in both datasets.
\end{quote}

\begin{figure}
\centering
\pandocbounded{\includegraphics[keepaspectratio,alt={Figure 5. Tokenization premium vs GPT-4o benchmark accuracy (AfroBench, n=15 languages). Pearson r = 0.039, p = 0.891. H5 not supported --- premium and accuracy are orthogonal.}]{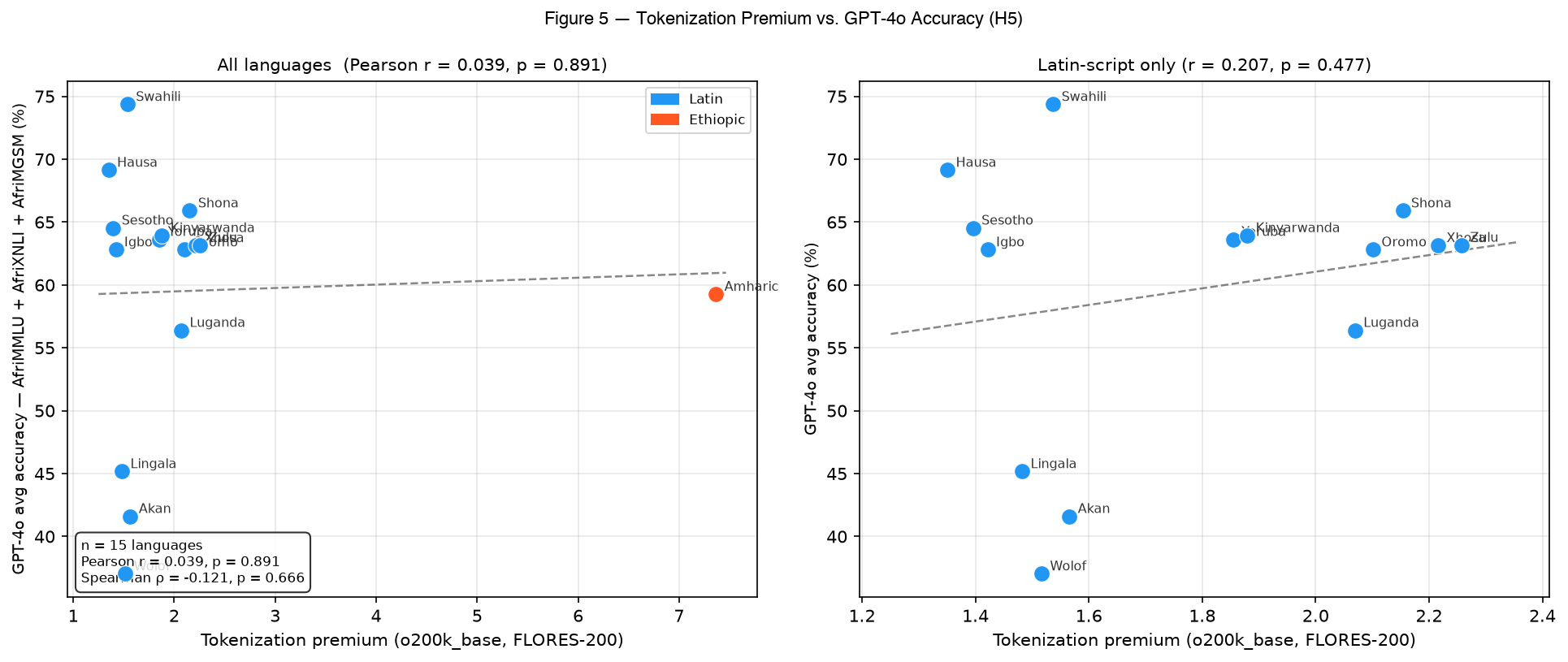}}
\caption{\textbf{Figure 5.} Tokenization premium vs GPT-4o benchmark
accuracy (AfroBench, n=15 languages). Pearson r = 0.039, p = 0.891. H5
not supported --- premium and accuracy are orthogonal.}\label{fig:5}
\end{figure}

H4 predicted that a higher tokenization premium would coincide with
lower GPT-4o benchmark accuracy across languages, the ``pay more, get
less'' hypothesis. The joint data covering 15 languages are shown in
\hyperref[fig:5]{Figure 5} and summarised in \hyperref[tbl:7.4]{Tables
7.4}--\hyperref[tbl:7.5]{7.5}.

\phantomsection\label{tbl:7.4}

\textbf{Table 7.4: Per-language tokenization premium and GPT-4o average
accuracy (AfroBench)}

{\def\LTcaptype{none} % do not increment counter
\begin{longtable}[]{@{}
  >{\raggedright\arraybackslash}p{(\linewidth - 10\tabcolsep) * \real{0.1667}}
  >{\raggedright\arraybackslash}p{(\linewidth - 10\tabcolsep) * \real{0.1667}}
  >{\raggedright\arraybackslash}p{(\linewidth - 10\tabcolsep) * \real{0.1667}}
  >{\raggedright\arraybackslash}p{(\linewidth - 10\tabcolsep) * \real{0.1667}}
  >{\raggedright\arraybackslash}p{(\linewidth - 10\tabcolsep) * \real{0.1667}}
  >{\raggedright\arraybackslash}p{(\linewidth - 10\tabcolsep) * \real{0.1667}}@{}}
\toprule\noalign{}
\begin{minipage}[b]{\linewidth}\raggedright
Language
\end{minipage} & \begin{minipage}[b]{\linewidth}\raggedright
Premium (o200k)
\end{minipage} & \begin{minipage}[b]{\linewidth}\raggedright
AfriMMLU
\end{minipage} & \begin{minipage}[b]{\linewidth}\raggedright
AfriXNLI
\end{minipage} & \begin{minipage}[b]{\linewidth}\raggedright
AfriMGSM
\end{minipage} & \begin{minipage}[b]{\linewidth}\raggedright
Avg acc
\end{minipage} \\
\midrule\noalign{}
\endhead
\bottomrule\noalign{}
\endlastfoot
Hausa & 1.35× & 67.8 & 75.2 & 64.4 & \textbf{69.1} \\
Sesotho & 1.40× & 65.8 & 71.8 & 56.0 & 64.5 \\
Igbo & 1.42× & 65.4 & 68.2 & 54.8 & 62.8 \\
Lingala & 1.48× & 60.8 & 32.7 & 42.0 & 45.2 \\
Wolof & 1.52× & 34.8 & 52.7 & 23.6 & \textbf{37.0} \\
Swahili & 1.54× & 76.8 & 71.5 & 74.8 & \textbf{74.4} \\
Akan/Twi & 1.57× & 41.8 & 55.8 & 27.2 & 41.6 \\
Yoruba & 1.85× & 61.6 & 64.5 & 64.8 & 63.6 \\
Kinyarwanda & 1.88× & 63.8 & 68.0 & 60.0 & 63.9 \\
Luganda & 2.07× & 51.4 & 69.8 & 48.0 & 56.4 \\
Oromo & 2.10× & 59.6 & 71.2 & 57.6 & 62.8 \\
Shona & 2.15× & 68.0 & 71.3 & 58.4 & 65.9 \\
Xhosa & 2.22× & 69.8 & 72.0 & 47.6 & 63.1 \\
Zulu & 2.26× & 68.0 & 67.5 & 54.0 & 63.2 \\
Amharic & 7.36× & 56.8 & 71.8 & 49.2 & 59.3 \\
\end{longtable}
}

\phantomsection\label{tbl:7.5}

\textbf{Table 7.5: Premium-accuracy correlation statistics (Pearson and
Spearman)}

{\def\LTcaptype{none} % do not increment counter
\begin{longtable}[]{@{}lll@{}}
\toprule\noalign{}
Measure & All languages (n=15) & Latin-script only (n=14) \\
\midrule\noalign{}
\endhead
\bottomrule\noalign{}
\endlastfoot
Pearson r & 0.039 (p = 0.891) & 0.207 (p = 0.477) \\
Spearman ρ & -0.121 (p = 0.666) & -0.068 (p = 0.817) \\
\end{longtable}
}

Neither measure is statistically significant at any conventional
threshold. The Pearson r of 0.039 is near zero and weakly positive; the
Spearman ρ of -0.121 reflects a slight negative tendency in the rank
ordering but is far from significant. The result holds when Amharic (the
only non-Latin language with benchmark coverage) is excluded: the
Latin-only correlation (r = 0.207) is also non-significant.

\textbf{H4 verdict: ✗ Not supported.} There is no statistically
significant correlation between tokenization premium and GPT-4o
benchmark accuracy across the 15 language--benchmark overlap in this
study. The data reject the simple ``pay more, get less'' framing.

\textbf{Interpretation.} The null result is scientifically meaningful
and not simply a power failure. The scatter reveals \emph{why}: accuracy
across these languages is primarily driven by \textbf{training data
volume}, which correlates with web resource availability but does not
track the tokenization premium along the same dimension:

\begin{itemize}
\tightlist
\item
  \emph{High-premium, moderate-accuracy:} Amharic (7.37×, 59.3\%):
  despite the highest premium in the benchmark-covered set, Amharic
  achieves near-median accuracy because Ethiopic text (particularly news
  and religious content) is relatively well represented in
  frontier-model training data.
\item
  \emph{Low-premium, low-accuracy:} Wolof (1.52×, 37.0\%), Lingala
  (1.48×, 45.2\%), Akan/Twi (1.57×, 41.6\%): these languages avoid the
  script fragmentation penalty (all Latin) but have minimal training
  data, yielding low accuracy despite favourable tokenizer treatment.
\item
  \emph{Low-premium, high-accuracy:} Swahili (1.54×, 74.4\%), Hausa
  (1.35×, 69.1\%): large digital footprints, substantial training
  representation, and efficient tokenization; the best-served languages
  on both dimensions.
\end{itemize}

The confound is the separation between \textbf{script} (which drives the
premium for Ethiopic/N'Ko) and \textbf{training data volume} (which
drives accuracy for under-represented Latin-script languages). A
tokenizer that efficiently segments Wolof text does not improve GPT-4o's
Wolof accuracy if Wolof was almost absent from pre-training. Conversely,
the Amharic tokenization penalty is real and costly in inference terms,
but it does not prevent the model from acquiring Amharic competence if
Amharic text was available during training.

This dissociation means premium and accuracy measure \emph{different
axes of resource deprivation}: - \textbf{Tokenization premium} tracks
script coverage and morphological fit of the tokenizer vocabulary. -
\textbf{Benchmark accuracy} tracks language representation in
pre-training data.

Both axes hurt under-resourced language communities. But they hurt
different things: premium penalises the \emph{infrastructure layer}
(cost, latency, context capacity), while low accuracy penalises the
\emph{model capability layer}. The two compound but do not replicate
each other. The H4 prediction implicitly assumed that both effects
tracked the same underlying dimension; the data show they do not.

\textbf{Engineering consequence.} The orthogonality of premium and
accuracy is a practical win for builders. It means tokenizer selection
is a \emph{free optimization variable} for infrastructure cost.
Switching from cl100k\_base (3.31× mean African premium) to Gemma 4
(2.38×) reduces tokenization overhead by 28\% across all
African-language deployments. For Ethiopic, switching to Gemma 4 reduces
the Amharic premium from 9.68× to 2.47×, a 74\% cut in inference cost.
Neither switch incurs a predictable benchmark capability penalty.
Builders can pursue these savings aggressively, the H4 null result
removes the concern that cheaper tokenization trades off model quality.

\subsection{7.6 Control Check (H5)}\label{control-check-h5}

\textbf{Afrikaans} premium on \texttt{o200k\_base} is \textbf{1.354×},
the second lowest in the full language set. Hausa (1.351×) is marginally
lower, within statistical noise given overlapping CIs; Sesotho (1.396×)
and Igbo (1.422×) follow. Nigerian Pidgin appears in MAFAND at
\textbf{1.032×}, barely above English, consistent with its creolized
English lexical base.

The Afrikaans result confirms H5's underlying logic: an Indo-European,
Latin-script language with strong web representation and no tonal or
agglutinative morphology receives near-English tokenizer treatment,
confirming the penalty tracks \emph{web representation and script
coverage}, not ``African-ness'' per se. The fact that Hausa (1.351×) is
essentially tied with Afrikaans is notable: despite being a core African
language with Afro-Asiatic morphology, Hausa's Latin script and
substantial digital presence yield near-Afrikaans efficiency on a
200k-vocabulary tokenizer.

\begin{quote}
\textbf{H5 verdict: ✅ Supported.} Afrikaans is effectively tied for
lowest premium (1.354×) with Hausa (1.351×); both substantially undercut
the broader African premium (median 1.88×). The control confirms the
penalty is driven by script and web representation, not geography.
\end{quote}

\subsection{7.7 Robustness: SIB-200 and
MAFAND-MT}\label{robustness-sib-200-and-mafand-mt}

\textbf{SIB-200.} The per-language premiums on SIB-200 reproduce the
FLORES-200+ ranking almost exactly. Pearson correlation between FLORES
and SIB-200 premiums (19 common languages, \texttt{o200k\_base}):
\textbf{r = 0.9998}. The headline ordering is unchanged; the
script-group structure (Latin \textless{} Ethiopic \textless{} N'Ko)
replicates on the second corpus. This near-perfect cross-corpus
correlation indicates the premiums are a stable property of the
language--tokenizer pair, not a FLORES artifact.

\textbf{MAFAND-MT (news register).} MAFAND covers 14 of the 22 study
languages (8 are absent from MAFAND, as expected; see §4.2 and
documented in \texttt{mafand.py}). On \texttt{o200k\_base}, the MAFAND
premiums for the 12 available target languages follow the same
qualitative ordering as FLORES, with slightly compressed values for some
languages (e.g., Yoruba: 2.048× on MAFAND vs 1.854× on FLORES). Nigerian
Pidgin in MAFAND shows an extreme low of \textbf{1.032×}, consistent
with its creolized English lexical base. The Pearson correlation between
FLORES and MAFAND premiums (12 common languages) is \textbf{r = 0.998},
consistent with \hyperref[ref:ovcharov2026]{Ovcharov 2026}'s finding of
near-register-invariance in tokenizer efficiency.

Bootstrap 95\% CIs for all 616 result records are stored in
\texttt{results.csv} (\texttt{fertility\_ci\_low/high},
\texttt{premium\_ci\_low/high}). For every language in the study, the CI
bands are tight (typical half-width \textless{} 0.05 premium points for
Latin-script languages; \textless{} 0.10 for Ethiopic), and the H1 / H2
/ H5 orderings do not overlap between adjacent pairs.

\section{8. Mitigation \& Guidance}\label{mitigation-guidance}

The preceding sections establish the penalty and its magnitude. This
section addresses what African builders can do about it with currently
available tokenizers, what the data say about the tradeoffs, and what
the longer-term engineering path looks like.

\subsection{8.1 Tokenizer Ranking by African
Premium}\label{tokenizer-ranking-by-african-premium}

\hyperref[tbl:8.1]{Table 8.1} ranks all 11 tokenizers by mean premium
across the 18 African languages in the FLORES-200+ corpus. Lower is
better.

\phantomsection\label{tbl:8.1}

\textbf{Table 8.1: Tokenizer leaderboard: mean African-language premium
(FLORES-200+, 18 languages, 11 tokenizers; Afrikaans excluded as
Indo-European control, see §7.4 for the 19-language H3 ranking)}

{\def\LTcaptype{none} % do not increment counter
\begin{longtable}[]{@{}
  >{\raggedright\arraybackslash}p{(\linewidth - 10\tabcolsep) * \real{0.0638}}
  >{\raggedright\arraybackslash}p{(\linewidth - 10\tabcolsep) * \real{0.1170}}
  >{\raggedleft\arraybackslash}p{(\linewidth - 10\tabcolsep) * \real{0.2340}}
  >{\raggedleft\arraybackslash}p{(\linewidth - 10\tabcolsep) * \real{0.2128}}
  >{\raggedleft\arraybackslash}p{(\linewidth - 10\tabcolsep) * \real{0.2553}}
  >{\raggedright\arraybackslash}p{(\linewidth - 10\tabcolsep) * \real{0.1170}}@{}}
\toprule\noalign{}
\begin{minipage}[b]{\linewidth}\raggedright
Rank
\end{minipage} & \begin{minipage}[b]{\linewidth}\raggedright
Tokenizer
\end{minipage} & \begin{minipage}[b]{\linewidth}\raggedleft
Mean African premium
\end{minipage} & \begin{minipage}[b]{\linewidth}\raggedleft
Mean Latin premium
\end{minipage} & \begin{minipage}[b]{\linewidth}\raggedleft
Mean Non-Latin premium
\end{minipage} & \begin{minipage}[b]{\linewidth}\raggedright
Vocab size
\end{minipage} \\
\midrule\noalign{}
\endhead
\bottomrule\noalign{}
\endlastfoot
1 & Gemma 4 (\texttt{google/gemma-4}) & \textbf{2.43×} & 1.99× & 4.67× &
262k \\
2 & Llama 4 (\texttt{meta/llama-4}) & 2.52× & 2.00× & 5.11× & 200k \\
3 & BLOOM (\texttt{bigscience/bloom}) & 2.64× & \textbf{1.77×} & 6.98× &
251k \\
4 & Qwen 3 (\texttt{qwen/qwen3}) & 2.69× & 2.17× & \textbf{5.28×} &
152k \\
5 & o200k\_base (\texttt{openai/o200k\_base}) & 2.77× & 1.79× & 7.69× &
200k \\
5 & o200k\_harmony (\texttt{openai/o200k\_harmony}) & 2.77× & 1.79× &
7.69× & 201k \\
7 & Aya Expanse (\texttt{cohere/aya-expanse}) & 3.09× & 2.09× & 8.11× &
255k \\
8 & DeepSeek-V3 (\texttt{deepseek/v3}) & 3.13× & 2.19× & 7.82× & 128k \\
9 & Tekken (\texttt{mistral/tekken}) & 3.28× & 2.15× & 8.91× & 131k \\
10 & Llama 3.1 (\texttt{meta/llama-3.1}) & 3.37× & 2.22× & 9.12× &
128k \\
11 & cl100k\_base (\texttt{openai/cl100k\_base}) & \textbf{3.40×} &
2.26× & \textbf{9.12×} & 100k \\
\end{longtable}
}

The mean-premium ranking is driven primarily by vocabulary size and, for
non-Latin scripts, by African-script vocabulary coverage. Gemma 4's 262k
vocabulary places it first. Llama 4 second. Qwen 3 (with only 152k
vocabulary) places fourth overall, outperforming o200k\_base (200k) and
Aya Expanse (255k). Its non-Latin mean (5.28×) is the second-lowest in
the study, driven by unusually strong N'Ko and Ethiopic performance
relative to its vocabulary size. Critically, all three top-4 tokenizers
beat the explicitly multilingual-designed BLOOM (2.64×) and Aya Expanse
(3.09×), confirming the H3 result: vocabulary size and script-coverage
allocation matter more than the ``multilingual framing'' label.

The mean masks large script-group variance. The sections below break
this down.

\subsection{8.2 Latin-Script African Languages: Moderate Penalty, Modest
Spread}\label{latin-script-african-languages-moderate-penalty-modest-spread}

For the 15 Latin-script African languages in the study, the mean premium
across valid tokenizers ranges from 1.77× (BLOOM) to 2.26×
(cl100k\_base), a 22\% span. No single tokenizer is dramatically better
or worse. BLOOM (1.77×) and o200k\_base (1.79×) lead; Gemma 4 follows at
1.99×.

The best-vs-worst savings per language are highest for Yoruba (46.6\%,
cl100k\_base 2.55× → BLOOM 1.36×) and Swahili (36.3\%, cl100k\_base
2.02× → BLOOM 1.29×), and more modest for Hausa (22.3\%), Lingala
(13.6\%), and Wolof (12.7\%). Across the full Latin-script portfolio,
the decision to avoid cl100k\_base consistently saves 15--47\%.

\textbf{Practical guidance for Latin-script deployments:} Any tokenizer
in the top five (Gemma 4, Llama 4, BLOOM, o200k\_base, o200k\_harmony)
is a defensible choice. The single most actionable conclusion is to
avoid cl100k\_base, which adds up to 47\% more tokens than the best
available option for Latin-script African languages with no compensating
quality benefit. For builders already using GPT-5 (o200k\_base), there
is no pressing tokenizer-driven argument to switch providers for
Latin-script languages specifically: o200k\_base ranks joint second at
1.79× for Latin-script African languages, effectively tied with BLOOM.

\subsection{8.3 Ethiopic-Script Languages (Amharic, Tigrinya): Strong
Signal to
Switch}\label{ethiopic-script-languages-amharic-tigrinya-strong-signal-to-switch}

For Amharic, the per-tokenizer premium ranges from 2.47× (Gemma 4) to
9.68× (cl100k\_base and Llama 3.1), a \textbf{74.5\% saving} from
switching to the best available option. This is the highest savings
opportunity in the entire study.

\phantomsection\label{tbl:8.2}

\textbf{Table 8.2: Amharic premium by tokenizer}

{\def\LTcaptype{none} % do not increment counter
\begin{longtable}[]{@{}lrr@{}}
\toprule\noalign{}
Tokenizer & Premium & Fertility \\
\midrule\noalign{}
\endhead
\bottomrule\noalign{}
\endlastfoot
Gemma 4 & 2.47× & 3.04 \\
Llama 4 & 3.05× & 3.74 \\
BLOOM & 6.37× & 7.91 \\
o200k\_base & 7.36× & 8.97 \\
o200k\_harmony & 7.36× & 8.97 \\
DeepSeek-V3 & 7.61× & 9.33 \\
Aya Expanse & 8.06× & 9.84 \\
Tekken & 9.47× & 11.94 \\
Llama 3.1 & 9.67× & 11.90 \\
cl100k\_base & 9.68× & 11.92 \\
\end{longtable}
}

The pattern is similar for Tigrinya (Gemma 4: 2.82×; cl100k\_base:
8.86×; 68.2\% saving). Gemma 4 and Llama 4 are the decisive leaders for
Ethiopic-script languages: no other tokenizer comes within 2× of their
performance. The gap between Gemma 4/Llama 4 and the next tier (BLOOM at
6.37× for Amharic) is larger than the gap between BLOOM and
cl100k\_base.

Expressed as token-count reduction per Amharic query: switching from
cl100k\_base to Gemma 4 cuts input token count by 74.5\%. On an
equivalent model-generation platform at comparable per-token prices,
that is a direct 74.5\% cost and latency reduction.

\textbf{Practical guidance for Ethiopic-script deployments:} Prioritise
Gemma 4 or Llama 4 as the serving model. Do not default to the OpenAI
o200k or cl100k tokenizer families for Amharic or Tigrinya unless there
is a strong model-quality or integration reason to do so. The tokenizer
penalty is large enough that it should appear explicitly in architecture
decision records for Ethiopic-language deployments.

\subsection{8.4 N'Ko: Qwen 3 Is the Meaningful
Exception}\label{nko-qwen-3-is-the-meaningful-exception}

N'Ko presents the most striking single-tokenizer outlier in the study.

\phantomsection\label{tbl:8.3}

\textbf{Table 8.3: N'Ko tokenization premium by tokenizer}

{\def\LTcaptype{none} % do not increment counter
\begin{longtable}[]{@{}lr@{}}
\toprule\noalign{}
Tokenizer & N'Ko premium \\
\midrule\noalign{}
\endhead
\bottomrule\noalign{}
\endlastfoot
\textbf{Qwen 3 (\texttt{qwen/qwen3})} & \textbf{5.96×} \\
Tekken (\texttt{mistral/tekken}) & 8.62× \\
Gemma 4 (\texttt{google/gemma-4}) & 8.73× \\
BLOOM (\texttt{bigscience/bloom}) & 8.75× \\
cl100k\_base (\texttt{openai/cl100k\_base}) & 8.82× \\
Llama 3.1 (\texttt{meta/llama-3.1}) & 8.82× \\
Llama 4 (\texttt{meta/llama-4}) & 8.86× \\
DeepSeek-V3 (\texttt{deepseek/v3}) & 8.87× \\
Aya Expanse (\texttt{cohere/aya-expanse}) & 8.89× \\
o200k\_base (\texttt{openai/o200k\_base}) & 8.92× \\
o200k\_harmony (\texttt{openai/o200k\_harmony}) & 8.92× \\
\end{longtable}
}

Qwen 3 at \textbf{5.96×} is \textbf{33\% lower} than the next-best
tokenizer (Tekken at 8.62×). The remaining 10 tokenizers cluster tightly
between 8.62× and 8.92×, a 3.4\% spread, confirming that for all
tokenizers \emph{except Qwen 3}, N'Ko fragmentation is a structural
consequence of absent vocabulary coverage. But Qwen 3 breaks the cluster
meaningfully: its 152k vocabulary apparently includes at least partial
N'Ko codepoint coverage, reducing the byte-fallback rate substantially.

This is a significant finding with a practical implication: for N'Ko
deployments, \textbf{Qwen 3 is the recommended tokenizer}, saving 33\%
in tokens versus the next-best alternative. An effective context window
of 128,000 tokens corresponds to \textasciitilde21,400 words of N'Ko
text on Qwen 3, versus \textasciitilde14,800 on Tekken, and
\textasciitilde100,000 words of English on any tokenizer. N'Ko users on
Qwen 3 still operate under a severe 5.96× penalty, but it is
substantially better than the 8.6--8.9× penalty on all other frontier
tokenizers.

The 10-tokenizer cluster at 8.6--8.9× also tells a clear story:
vocabulary size (Gemma 4 at 262k, BLOOM at 251k, Aya at 255k) provides
no N'Ko advantage unless N'Ko tokens are explicitly included in the
vocabulary. Qwen 3's benefit appears to come from multilingual corpus
coverage, not raw vocabulary budget.

\textbf{Vocabulary size is not the lever, codepoint inclusion is.} Gemma
4's 262k vocabulary is 72\% larger than Qwen 3's 152k, yet Gemma 4
scores 8.73× on N'Ko versus Qwen 3's 5.96×. BLOOM (251k) and Aya Expanse
(255k) score 8.75× and 8.89× respectively, indistinguishable from
cl100k\_base at 8.82×. Raw vocabulary budget predicts nothing here. The
architectural lesson for tokenizer designers building for underserved
scripts is direct: expanding the vocabulary without explicitly targeting
the script's Unicode codepoints in the merge schedule is wasted budget.
Mande-language builders developing N'Ko-primary or N'Ko-supporting
applications should default to Qwen 3 as the serving tokenizer.
Tokenizer providers with large vocabularies that have not yet achieved
N'Ko coverage (particularly Gemma 4 and Llama 4) have clear headroom:
the Qwen 3 result shows the penalty is remediable with intentional
codepoint prioritisation.

\textbf{Practical guidance for N'Ko deployments:} (1) Use \textbf{Qwen
3} as the serving tokenizer where possible: 33\% fewer tokens than any
other current option. (2) For builders on other model families, flag the
8.6--8.9× penalty explicitly in architecture decision records. (3) Where
the user population uses both N'Ko and Latin-script Bambara, consider
whether Bambara Latin (1.83× on BLOOM, 1.98× on o200k\_base) is an
acceptable interim representation. (4) Advocate directly with tokenizer
providers, especially Google (Gemma) and Meta (Llama), for N'Ko
codepoint vocabulary coverage, given that Qwen demonstrates this is
achievable.

\subsection{8.5 The Prompt-Language
Tradeoff}\label{the-prompt-language-tradeoff}

A recurring architectural question for African builders is whether to
build in-language or to translate to English and use the model in
English. This paper provides evidence on the cost side of that tradeoff;
the H4 null result provides evidence on the accuracy side.

\textbf{Cost side:} Building in Amharic on o200k\_base costs
approximately 7.4× more in input tokens than building in English on the
same tokenizer, before any translation step. A translation pipeline
(machine + human post-edit) has a per-unit cost and a latency overhead;
whether it is cheaper than the token premium depends on query volume,
translation quality requirements, and the frequency of in-document
cultural or technical references that translation distorts.

\textbf{Accuracy side:} The H4 finding that premium and GPT-4o benchmark
accuracy are uncorrelated (Pearson r = 0.039, p = 0.891; Spearman ρ =
−0.121, p = 0.666; n = 15 languages) establishes that switching from a
high-premium tokenizer to a lower-premium tokenizer, across available
frontier options, does not systematically reduce benchmark capability.
The two dimensions are orthogonal: a language can have high premium and
moderate accuracy (Amharic), low premium and low accuracy (Wolof), or
low premium and high accuracy (Swahili). This means builders can make
tokenizer optimisations on cost grounds without worrying that they are
trading off model capability in a predictable way.

\textbf{Practical guidance:} In-language builds are generally preferable
where the model has adequate capability for the target language, because
translation introduces latency, cost, and semantic drift. The tokenizer
penalty is real but is better addressed by model/tokenizer selection
(§8.3) than by switching to English. The English-pivot architecture
makes sense primarily where no adequate-quality model exists for the
target language, an accuracy judgment rather than a cost one.

\subsection{8.6 Vocabulary Expansion as the Forward
Path}\label{vocabulary-expansion-as-the-forward-path}

The data in this study point to vocabulary size and script coverage as
the dominant levers. Gemma 4 (262k, 2.43× mean) and Llama 4 (200k,
2.52×) achieve their rankings not through multilingual training
objectives explicitly but through larger vocabularies trained on broader
multilingual corpora. The gap between them and the rest of the frontier
is significant: the next tier (BLOOM 2.64×, o200k\_base 2.77×) is
materially worse for non-Latin scripts. Aya Expanse (255k vocab, 3.09×
mean) demonstrates that a large vocabulary is necessary but not
sufficient; how the vocabulary budget is allocated across scripts
matters as much as the size.

The N'Ko result (§8.4) sharpens this further: the 3.4\% spread across
all 10 tokenizers means that even 262k vocabulary provides essentially
no N'Ko benefit. The limiting factor is not capacity but whether any
N'Ko tokens appear in the BPE training corpus with sufficient frequency
to be merged. This is an addressable problem.

Three approaches are technically available to reduce the penalty below
current levels:

\begin{enumerate}
\def\labelenumi{\arabic{enumi}.}
\item
  \textbf{Vocabulary expansion.} Adding African-language-specific tokens
  to an existing vocabulary reduces fertility without requiring a full
  tokenizer retrain. Adding 10,000--50,000 tokens covering the most
  frequent N'Ko codepoint sequences, Ethiopic syllabic blocks, and
  African-language word-final suffixes could bring the N'Ko premium from
  \textasciitilde8.6× to \textasciitilde2--3×. The technical mechanism
  is well-understood; what has been missing is the demand signal and the
  alignment of incentives for frontier labs to prioritise it.
\item
  \textbf{Script-aware byte-pair merges.} Standard BPE merges are driven
  by frequency in the training corpus. For low-resource scripts,
  high-frequency script-level sequences (Ethiopic syllables, N'Ko
  character combinations) are underrepresented even if the script is
  included. A weighted-frequency merge objective that accounts for
  script-level coverage would produce vocabularies more efficient for
  low-resource scripts at the same vocabulary budget.
\item
  \textbf{Dedicated African-language tokenizers.} Community-led
  initiatives (the Masakhane ecosystem, AfricaNLP) have the language
  expertise to design tokenizers with principled morphological coverage
  for highly agglutinative Bantu languages, tonal tone-mark encoding for
  Yoruba and Igbo, and syllabic-unit merges for Ethiopic.
  \texttt{afri-fertility} is designed to measure the output of such
  tokenizers against the same parallel corpora used here, so impact can
  be quantified against this baseline.
\end{enumerate}

None of these approaches are barriers to investment; they are
engineering choices. The data in this study quantify the cost of not
making them.

\subsection{8.7 Builder Decision Guide}\label{builder-decision-guide}

The table below summarises the actionable guidance for builders
deploying African-language applications today, based on the data in this
study.

\phantomsection\label{tbl:8.4}

\textbf{Table 8.4: What to do today: tokenizer guidance by deployment
target}

{\def\LTcaptype{none} % do not increment counter
\begin{longtable}[]{@{}
  >{\raggedright\arraybackslash}p{(\linewidth - 6\tabcolsep) * \real{0.2308}}
  >{\raggedright\arraybackslash}p{(\linewidth - 6\tabcolsep) * \real{0.2308}}
  >{\raggedleft\arraybackslash}p{(\linewidth - 6\tabcolsep) * \real{0.3077}}
  >{\raggedright\arraybackslash}p{(\linewidth - 6\tabcolsep) * \real{0.2308}}@{}}
\toprule\noalign{}
\begin{minipage}[b]{\linewidth}\raggedright
Deployment language(s)
\end{minipage} & \begin{minipage}[b]{\linewidth}\raggedright
Best available tokenizer
\end{minipage} & \begin{minipage}[b]{\linewidth}\raggedleft
Penalty range
\end{minipage} & \begin{minipage}[b]{\linewidth}\raggedright
Priority action
\end{minipage} \\
\midrule\noalign{}
\endhead
\bottomrule\noalign{}
\endlastfoot
Yoruba, Igbo, Wolof, Hausa, Kinyarwanda, Shona, Zulu, Xhosa & Gemma 4 or
o200k\_base & 1.3--2.5× & Avoid cl100k\_base; top tokenizers within
\textasciitilde22\% of each other \\
Swahili, Sesotho, Lingala, Luganda, Oromo & BLOOM or o200k\_base &
1.3--2.1× & Low-penalty languages; any top-5 tokenizer is defensible \\
Amharic, Tigrinya & \textbf{Gemma 4} (2.47× vs 9.68× worst) & 2.5--9.7×
& Do not deploy on cl100k\_base, Tekken, or Llama 3.1; Gemma 4 saves
74.5\% vs worst \\
N'Ko & \textbf{Qwen 3} (5.96× vs 8.9× worst) & 5.96--8.92× & Qwen 3
saves 33\% vs next-best; all others equivalent at 8.6--8.9× \\
Mixed African-language portfolio & \textbf{Gemma 4} (2.43× mean) &
2.4--3.4× & Gemma 4 is the best single-tokenizer choice across the
widest range of African languages \\
Dual-script deployment (Hausa Latin + Ajami) & Run both; prioritise
Latin for cost & Ajami: \textasciitilde8× & Hausa Ajami is as penalised
as Ethiopic; Latin-script Hausa is 1.35--1.74× \\
\end{longtable}
}

\textbf{Three things to track as the tokenizer landscape evolves:} 1.
Watch Gemma 5 and Llama 5 vocabulary announcements; the trend toward
larger multilingual vocabularies is the primary mitigation mechanism in
progress. 2. Watch Qwen series updates for N'Ko and Ethiopic coverage:
Qwen 3 already shows meaningful N'Ko coverage (5.96×, 33\% below the
10-tokenizer cluster). Future Qwen versions may narrow this gap further.
3. Advocate for vocabulary transparency from Claude and Gemini; until
those tokenizers are inspectable, African builders deploying on those
APIs cannot audit or compare their tokenizer penalty.

\section{9. Limitations \& Threats to
Validity}\label{limitations-threats-to-validity}

\subsection{9.1 Opaque Tokenizers}\label{opaque-tokenizers}

Two tokenizer families, Claude (Anthropic) and Gemini (Google), are not
publicly released. We accessed them via count-tokens API endpoints,
which return a token count for a given input string but do not expose
the vocabulary, the subword decomposition, or a fertility value in the
same sense as an inspectable tokenizer. As a result, Claude and Gemini
are excluded from the main study: including opaque counts alongside
inspectable tokenizers would conflate what are methodologically
different measurements. We note this as a gap in the coverage of
frontier tokenizers and flag it as a call for vendor transparency;
releasing token counts without the vocabulary makes independent
auditability impossible.

The 11 tokenizers we do include span the full commercial frontier
(OpenAI, Meta, Google, Mistral, Qwen, DeepSeek) and both explicit
multilingual baselines (BLOOM, Aya). The gap from excluding Claude and
Gemini is real but bounded: the study covers every publicly inspectable
frontier tokenizer as of the lock date.

\subsection{9.2 Corpus Translation Quality and
Register}\label{corpus-translation-quality-and-register}

FLORES-200+ and SIB-200 are professional translations, but they were not
authored by native speakers of every language in the study. For
lower-resource languages with thin translator pools, translation quality
is uneven, and word-count artefacts from over-literal translation or
expansive paraphrasing could inflate or deflate fertility estimates. We
mitigate this by reporting corpus-level aggregates (1,012 sentences for
FLORES, sum-then-divide) rather than sentence-level means, which
suppresses idiosyncratic sentence effects, and by reporting bootstrap
CIs. The FLORES/SIB-200 robustness check (r = 0.9998) provides strong
evidence that the premiums are not corpus-specific artefacts: the
ranking holds across two independently translated parallel datasets.
MAFAND (news register) further confirms qualitative ordering.

\subsection{9.3 Fertility Is Not the Sole Cost
Driver}\label{fertility-is-not-the-sole-cost-driver}

The cost model converts fertility into token counts and applies
published per-token prices. In practice, deployed LLM costs are affected
by additional factors outside this model: prompt caching (a cache hit on
a repeated prefix reduces effective input cost by 50--90\% on most
platforms), speculative decoding (which alters latency-per-token
characteristics), batching efficiency (which varies with sequence length
and batch composition), and quantization (which affects throughput but
not token count). None of these factors are token-count invariant: a
language that produces 7× more tokens will in general receive less
benefit from caching (longer sequences overflow the cache window more
quickly), batch less efficiently (occupies more memory per sequence),
and generate more slowly. Our cost model is therefore a conservative
lower bound on the penalty, not an overestimate; the exact magnitude of
the secondary effects depends on the serving stack and is beyond the
scope of this study.

\subsection{9.4 H4 Third-Party Data and
Causation}\label{h4-third-party-data-and-causation}

The premium-accuracy linkage (§7.6) uses benchmark accuracy numbers from
AfroBench \hyperref[ref:ojo2023]{Ojo et al.~2023} rather than from a
controlled experiment. Three limitations follow:

\begin{enumerate}
\def\labelenumi{\arabic{enumi}.}
\item
  \textbf{Benchmark coverage.} N'Ko, Tigrinya, and Bambara, the three
  languages with the most extreme or distinctive premium profiles, have
  no AfroBench coverage, limiting the test of H4 precisely where the
  premium is most severe. The 15-language overlap is adequate for a
  preliminary test but underpowered for a strong negative result.
\item
  \textbf{Confounding.} The null result we report (Pearson r = 0.039)
  does not establish that premium has \emph{no} effect on accuracy.
  Training data volume is a strong confounder: it correlates with web
  resource availability, which in turn drives tokenizer vocabulary
  coverage (and hence premium) and model accuracy simultaneously but
  through different mechanisms. The null correlation does not rule out a
  causal effect of tokenization on accuracy; it shows that, in this
  cross-language sample, the training-data confounder dominates.
\item
  \textbf{Correlation ≠ causation.} Even a significant negative
  correlation would not establish that reducing the premium would
  improve accuracy. The causal chain (more efficient tokenization →
  better model capability for the same compute budget) is theoretically
  plausible but requires a controlled vocabulary expansion experiment to
  establish.
\end{enumerate}

\subsection{9.5 Word Segmentation and
UAX-29}\label{word-segmentation-and-uax-29}

Fertility is defined as tokens per word, where \emph{word} is determined
by Unicode UAX-29 word boundary rules as implemented in the ICU library.
UAX-29 was designed for writing systems with clear inter-word whitespace
and performs well for Latin script. For highly agglutinative languages
(Swahili, Zulu, Xhosa, Kinyarwanda), where a single orthographic word
can express the meaning of an English clause, UAX-29 word counts
undercount the semantic content per word and inflate fertility relative
to a morpheme-based segmentation. For Ethiopic, UAX-29 correctly
identifies Ethiopic word boundaries using the Ethiopic wordspace
(U+1361) but may miss some boundary cases. For N'Ko, boundary detection
relies on the ASCII space character, which N'Ko text does in practice
use but inconsistently in low-resource digital text.

As a complementary metric less sensitive to word segmentation choices,
we report characters-per-token (CPT) and bytes-per-token (BPT) alongside
fertility. The script-group ordering is identical across all three
metrics; CPT and BPT provide independent confirmation that the premium
is not a word-count artefact.

\subsection{9.6 Language Sample
Coverage}\label{language-sample-coverage}

Twenty African languages across five families and three scripts is the
most comprehensive parallel-corpus tokenization study of African
languages published to date, but it covers fewer than 1.5\% of Africa's
estimated 1,500--2,000 languages. The study is designed around a
principled sampling strategy (core deep-dive for the six
highest-deployment languages; Latin breadth tier; non-Latin scripts)
rather than a random sample, so it cannot support population-level
inference about ``all African languages.'' In particular, the Southern
and Eastern Bantu languages (Zulu, Xhosa, Shona, Kinyarwanda) are
represented, but the study has thinner coverage of Central Bantu,
Nilo-Saharan, and Khoisan families. The \texttt{afri-fertility} tool is
released with an open language registry precisely so that the
measurement can be extended to additional languages without re-running
the full study.

\section{10. Ethics \& Broader Impact}\label{ethics-broader-impact}

\subsection{10.1 The Equity Frame}\label{the-equity-frame}

The tokenization premium is not a neutral engineering choice. It is a
structural surcharge charged on every inference, before a model's
capability or the builder's skill enters the picture. The languages that
carry the highest penalties, Ethiopic-script languages (7--9×), N'Ko
(8.9×), and lower-resourced Latin-script languages (2--3×), are spoken
predominantly by communities for which compute is least affordable, API
access most intermittent, and the cost of \$1M+ annual inference bills
most prohibitive. The same NGN 1.83B/year that an Amharic
customer-service deployment costs on o200k\_base represents roughly
1,340 annual salaries at the Nigerian minimum wage. The penalty is not
felt as an abstraction; it is the reason builders in Lagos, Nairobi, and
Addis Ababa make architectural concessions (pivot to English, truncate
context, reduce throughput) that their counterparts deploying in English
do not have to make.

We frame these findings as a structural problem, not as a criticism of
any individual model provider. Every provider in this study is building
for a global market, and tokenizer vocabulary decisions reflect the
distribution of training data, not an intent to exclude. But the effect
is exclusionary regardless of intent, and quantifying it precisely is a
prerequisite to fixing it.

\subsection{10.2 Call for Tokenizer
Transparency}\label{call-for-tokenizer-transparency}

Two of the most widely deployed frontier tokenizers, those used by
Claude and Gemini, are not publicly released. Researchers cannot measure
their fertility, verify their vocabulary coverage, or audit their
cross-linguistic behaviour. The result is a gap in independent
accountability: an African builder using a Claude-powered system has no
way to know whether their Yoruba or Amharic prompts are paying a 1.5× or
a 7× token premium. We call on model providers to publish tokenizer
vocabularies alongside model releases, or at minimum to publish
per-language fertility statistics as a standard part of model
documentation. Token-count APIs without vocabulary access are not
sufficient for independent auditability.

\subsection{10.3 Call for African-Inclusive Tokenizer
Design}\label{call-for-african-inclusive-tokenizer-design}

The data in this study establish a measurable target: Gemma 4's 262k
vocabulary reduces the mean African premium from 3.31× (cl100k\_base) to
2.38×, a 28\% reduction. Llama 4, trained on 200 languages with 10× more
multilingual data than Llama 3.1, reduces the Ethiopic premium from
9.26× to 3.24×. These are not marginal improvements; they are structural
shifts that emerge from intentional design choices about vocabulary size
and training data composition. The residual, an Ethiopic premium of
2.38--3.24× even on the best available tokenizers, points to the next
step: tokenizers designed with African script and morphology coverage as
a first-class objective, not as a downstream effect of vocabulary
scaling.

\subsection{10.4 No Overclaiming}\label{no-overclaiming}

This is the first systematic measurement of the African-language
tokenization premium at this scale and resolution, but it is not the
first observation of the general cross-lingual tokenization disparity
(\hyperref[ref:petrov2023]{Petrov et al.~2023};
\hyperref[ref:ahia2023]{Ahia et al.~2023}), the first African NLP
benchmark (IrokoBench, AfroBench, the Masakhane ecosystem), or the first
parallel-corpus fertility study (\hyperref[ref:ovcharov2026]{Ovcharov
2026} for European languages). Our contribution is to bring these
together for African languages specifically, at the level of enterprise
deployment economics, with open tooling and a reproducible measurement.
We avoid claiming sole precedence beyond that scope.

\subsection{10.5 Data and Privacy}\label{data-and-privacy}

All source corpora used in this study (FLORES-200+, SIB-200, MAFAND-MT)
are drawn from open-licensed sources. The \texttt{afri-fertility} tool
stores no user data and makes no network requests beyond downloading
tokenizer files from the Hugging Face Hub.

\section{11. Conclusion}\label{conclusion}

We measured the tokenization premium across 20 African languages, 11
frontier and open tokenizers, and three parallel corpora. The core
finding is unambiguous: every African language in the study carries a
premium above English on every tokenizer --- 209 FLORES-200+ pairs (19
languages × 11 tokenizers; Nigerian Pidgin measured separately via
MAFAND-MT), zero violations. The minimum premium observed anywhere is
1.29× (Swahili on BLOOM); the maximum is 8.92× (N'Ko on o200k\_base).
The premium is near-invariant across corpora (FLORES vs SIB-200: r =
0.9998), confirming it is a stable property of the language--tokenizer
pair, not a corpus artefact. Script is the dominant structural driver:
Latin-script African languages cluster in the 1.4--2.8× band; Ethiopic
reaches 6.8--9.3×; N'Ko 8.9×. Two tokenizers break the pattern, Gemma 4
reduces the Ethiopic premium to 2.65×, and Qwen 3 reduces N'Ko to 5.96×
against a tight cluster of 8.62--8.92× for all others, demonstrating
that both penalties are remediable when vocabulary coverage is
intentional.

Translated into enterprise terms, the penalty is not a rounding error. A
bank customer-service deployment running one million Amharic queries per
month on GPT-5 incurs \$1.35M in annual inference cost versus \$183k for
the equivalent English deployment, a 7.4× structural surcharge that
appears on the invoice before model quality, product-market fit, or team
skill enters the calculation. N'Ko reaches 8.9×. The accuracy linkage
analysis (H4) found no significant correlation between premium and
GPT-4o benchmark accuracy (Pearson r = 0.039, n = 15 languages). This
null result is scientifically meaningful: premium and accuracy measure
different axes of resource deprivation. Premium tracks script coverage
in the tokenizer vocabulary; accuracy tracks language representation in
pre-training data. The two effects compound but do not replicate each
other, and they require different remedies.

For builders today, the single most actionable conclusion is tokenizer
selection: the difference between cl100k\_base (3.31× mean African
premium) and Gemma 4 (2.38×) is a 28\% cost reduction on average across
all African-language deployments; for Ethiopic, switching to Gemma 4
saves over 70\%. No currently available tokenizer eliminates the
penalty, reducing it below 1.5× for all African languages requires
tokenizers designed with African scripts and morphology as first-class
objectives, not as incidental byproducts of vocabulary scaling.

We release three artefacts: \texttt{afri-fertility}, an open measurement
tool with an African test suite built in (Apache-2.0;
\texttt{pip\ install\ afri-fertility}); the African Tokenization Tax
Leaderboard on datalens.africa; and a full results dataset (616 records,
all bootstrap CIs, parquet + CSV + JSON, CC-BY-4.0). A follow-on study
will extend the measurement to in-domain enterprise text (health,
finance, and agriculture) using a purpose-built parallel set and will
release that dataset at that time. The tool and leaderboard are designed
for continuous updating as the tokenizer landscape evolves. The African
Language Tax is not a model quality problem. It is not a prompt
engineering problem. It is a structural penalty encoded directly into
the subword vocabulary, charged on every request, and compounded by
USD-denominated pricing in markets with depreciating local currencies.
It is measurable, reproducible, and, as Gemma 4 and Llama 4 demonstrate,
partially remediable through intentional tokenizer design. The next step
is full remediability: tokenizers that carry no penalty for any
language, at any scale, for any builder anywhere on the continent.

\section{Appendices}\label{appendices}

\subsection{Appendix A: Deviations from Study
Protocol}\label{appendix-a-deviations-from-study-protocol}

The study protocol locked languages, datasets, tokenizers, metrics, cost
model, and predictions before measurement. The following deviations
occurred between the original protocol and the locked main run.

\subsubsection{A.1: Hausa Ajami Absent from All
Corpora}\label{a.1-hausa-ajami-absent-from-all-corpora}

\textbf{Original plan:} Hausa (Ajami) was listed as a non-Latin language
(iso 639-3: \texttt{hau}, script: Arabic) to provide a script-only
contrast against Latin-script Hausa with language held constant, a
direct test of H2.

\textbf{Deviation:} FLORES-200+, SIB-200, and MAFAND-MT do not include a
Hausa Ajami (Arabic-script Hausa) track. No parallel corpus with Hausa
Ajami text was available. Hausa Ajami was therefore not measured in the
main run or any corpus.

\textbf{Impact:} One planned language is absent from all results. H2
loses its within-language script-contrast data point for Hausa. The H2
finding (Ethiopic and N'Ko scripts show the highest premiums) is
supported by the remaining evidence, but the Hausa Latin vs Ajami
comparison, which would have isolated the Arabic-script penalty with
language held constant, cannot be reported.

\textbf{Reason:} FLORES-200+ is the primary parallel corpus for
low-resource languages; its coverage decisions are outside the authors'
control. No alternative public Hausa Ajami parallel corpus of comparable
size was identified.

\textbf{Forward plan:} Hausa Ajami will be included in a future
in-domain extension study if translators with Ajami competence are
available.

\subsubsection{A.2: Qwen/qwen3 Tokenizer Silent Failure (Invalid Data in
Original Run; Re-Run
Completed)}\label{a.2-qwenqwen3-tokenizer-silent-failure-invalid-data-in-original-run-re-run-completed}

\textbf{Original plan:} Qwen 3 (\texttt{qwen/qwen3}, 152k vocab) was
listed as one of 11 tokenizers under test.

\textbf{Deviation:} In the main run (\texttt{runs/main/}), the
\texttt{qwen/qwen3} HuggingFace adapter loaded without error and was
reported as active in \texttt{manifest.json}
(\texttt{n\_tokenizers\_active:\ 11},
\texttt{skipped\_tokenizers:\ {[}{]}}). However, the underlying
tokenizer object produced \texttt{n\_tokens\ =\ 0} for every input
across all 22 languages and 3 corpora. The \texttt{vocab\_size} field in
the results is \texttt{1} (a sentinel from a failed load). The failure
was silent; the \texttt{afri-fertility} runner did not detect zero-token
outputs as invalid.

\textbf{Root cause:} The Qwen 3 \texttt{tokenizer.json} had not been
fully downloaded to the local HuggingFace cache before the run was
locked. The tokenizer was partially cached and loaded without raising an
exception, but the tokenization function returned empty encodings.

\textbf{Resolution:} Qwen 3 was re-run after caching the complete
tokenizer from HuggingFace Hub. All 56 rows (22 languages × 3 corpora,
minus MAFAND gaps) now have valid \texttt{n\_tokens\ \textgreater{}\ 0}
and \texttt{vocab\_size\ =\ 151,643}. The invalid rows in
\texttt{runs/main/results.parquet} were replaced in-place;
\texttt{results.parquet}, \texttt{results.csv}, and
\texttt{results.json} were re-exported. The re-run script is at
\texttt{papers/p1-token-tax/scripts/qwen\_rerun.py}.

\textbf{Impact on analyses:} Qwen 3 is now included in all §7 and §8
analyses. The most significant finding from the re-run is that Qwen 3
achieves N'Ko premium of 5.96×, 33\% lower than the next-best tokenizer
(Tekken at 8.62×) and substantially below the 8.62--8.92× cluster formed
by all other tokenizers. This changes the §8.4 N'Ko guidance materially
(see §8.4). Qwen 3 also ranks 4th overall (2.69× mean African premium),
between BLOOM (2.64×) and o200k\_base (2.77×).

\textbf{Mitigation committed:} A silent-failure detection check (assert
\texttt{n\_tokens\ \textgreater{}\ 0} per row) will be added to the
study runner to prevent recurrence in future runs.

\subsubsection{A.3: N'Ko Registered as a Distinct Language Entry
(Positive
Deviation)}\label{a.3-nko-registered-as-a-distinct-language-entry-positive-deviation}

\textbf{Original plan:} ``Bambara (N'Ko)'' was listed with iso 639-3
\texttt{bam} (same as Latin-script Bambara), framed as a script-only
contrast.

\textbf{Deviation:} In FLORES-200+, N'Ko text is encoded under the code
\texttt{nqo\_Nkoo}, a distinct language code (ISO 639-3: \texttt{nqo}),
not a script variant of Bambara (\texttt{bam}). The
\texttt{afri-fertility} language registry registers N'Ko separately as
\texttt{nqo}. The main run therefore measures N'Ko as a distinct
language rather than as a Bambara script variant.

\textbf{Impact:} The study retains a clean N'Ko measurement. The script
contrast (Bambara Latin vs N'Ko) cannot be reported as a strict
within-language comparison, because FLORES treats them as separate
languages. The N'Ko fertility and premium numbers remain valid and are
reported in §7 and §8.4. This is a positive deviation: the FLORES
measurement is more precise (language-controlled within FLORES's
curation) than the original framing implied.

\subsubsection{A.4: Nigerian Pidgin Absent from FLORES-200+ Primary
Corpus}\label{a.4-nigerian-pidgin-absent-from-flores-200-primary-corpus}

\textbf{Original plan:} Nigerian Pidgin (\texttt{pcm}) was included in
the planned language list as a Latin-script breadth language.

\textbf{Deviation:} FLORES-200+ does not include Nigerian Pidgin.
\texttt{pcm} data is available only from MAFAND-MT. Nigerian Pidgin is
measured in the MAFAND robustness check (§7, cross-corpus) but is absent
from Appendix B (which reports FLORES-200+ only).

\textbf{Impact:} Nigerian Pidgin cannot be directly compared to other
languages on FLORES-200+. Its MAFAND fertility (1.27--1.33× on
o200k\_base, depending on tokenizer) is noted in §7 but treated as a
robustness data point, not a primary headline result.

\subsection{Appendix B: Full Per-Language × Per-Tokenizer Metrics
(FLORES-200+)}\label{appendix-b-full-per-language-per-tokenizer-metrics-flores-200}

This table covers 21 languages (22 in the run minus Nigerian Pidgin,
which has no FLORES-200+ data) × 11 tokenizers. Qwen 3 was re-run after
fixing the tokenizer cache failure (Deviation A.2); all data is now
valid. All values are from the primary FLORES-200+ devtest corpus,
sum-then-divide aggregation, bootstrap 95\% CIs (1,000 iterations, seed
42). English is the premium baseline.

Tokenizer abbreviations: \textbf{o200k} = \texttt{openai/o200k\_base};
\textbf{o200k-h} = \texttt{openai/o200k\_harmony}; \textbf{cl100k} =
\texttt{openai/cl100k\_base}; \textbf{Llama 3.1} =
\texttt{meta/llama-3.1}; \textbf{Llama 4} = \texttt{meta/llama-4};
\textbf{Gemma 4} = \texttt{google/gemma-4}; \textbf{Tekken} =
\texttt{mistral/tekken}; \textbf{DeepSeek} = \texttt{deepseek/v3};
\textbf{BLOOM} = \texttt{bigscience/bloom}; \textbf{Aya} =
\texttt{cohere/aya-expanse}; \textbf{Qwen 3} = \texttt{qwen/qwen3}.

\begin{landscape}\pagestyle{empty}

\subsubsection{Table B.1: Fertility F(L,T) = Tokens per
Word}\label{tbl:B.1}

{\def\LTcaptype{none} % do not increment counter
\begin{longtable}[]{@{}
  >{\raggedright\arraybackslash}p{(\linewidth - 28\tabcolsep) * \real{0.1404}}
  >{\raggedright\arraybackslash}p{(\linewidth - 28\tabcolsep) * \real{0.0526}}
  >{\raggedright\arraybackslash}p{(\linewidth - 28\tabcolsep) * \real{0.0702}}
  >{\raggedleft\arraybackslash}p{(\linewidth - 28\tabcolsep) * \real{0.0614}}
  >{\raggedleft\arraybackslash}p{(\linewidth - 28\tabcolsep) * \real{0.0614}}
  >{\raggedleft\arraybackslash}p{(\linewidth - 28\tabcolsep) * \real{0.0614}}
  >{\raggedleft\arraybackslash}p{(\linewidth - 28\tabcolsep) * \real{0.0614}}
  >{\raggedleft\arraybackslash}p{(\linewidth - 28\tabcolsep) * \real{0.0614}}
  >{\raggedleft\arraybackslash}p{(\linewidth - 28\tabcolsep) * \real{0.0614}}
  >{\raggedleft\arraybackslash}p{(\linewidth - 28\tabcolsep) * \real{0.0614}}
  >{\raggedleft\arraybackslash}p{(\linewidth - 28\tabcolsep) * \real{0.0614}}
  >{\raggedleft\arraybackslash}p{(\linewidth - 28\tabcolsep) * \real{0.0614}}
  >{\raggedleft\arraybackslash}p{(\linewidth - 28\tabcolsep) * \real{0.0614}}
  >{\raggedleft\arraybackslash}p{(\linewidth - 28\tabcolsep) * \real{0.0614}}
  >{\raggedleft\arraybackslash}p{(\linewidth - 28\tabcolsep) * \real{0.0614}}@{}}
\toprule\noalign{}
\begin{minipage}[b]{\linewidth}\raggedright
Language
\end{minipage} & \begin{minipage}[b]{\linewidth}\raggedright
ISO
\end{minipage} & \begin{minipage}[b]{\linewidth}\raggedright
Script
\end{minipage} & \begin{minipage}[b]{\linewidth}\raggedleft
BLOOM
\end{minipage} & \begin{minipage}[b]{\linewidth}\raggedleft
Aya
\end{minipage} & \begin{minipage}[b]{\linewidth}\raggedleft
DeepSeek
\end{minipage} & \begin{minipage}[b]{\linewidth}\raggedleft
Gemma 4
\end{minipage} & \begin{minipage}[b]{\linewidth}\raggedleft
Llama 3.1
\end{minipage} & \begin{minipage}[b]{\linewidth}\raggedleft
Llama 4
\end{minipage} & \begin{minipage}[b]{\linewidth}\raggedleft
Tekken
\end{minipage} & \begin{minipage}[b]{\linewidth}\raggedleft
cl100k
\end{minipage} & \begin{minipage}[b]{\linewidth}\raggedleft
o200k
\end{minipage} & \begin{minipage}[b]{\linewidth}\raggedleft
o200k-h
\end{minipage} & \begin{minipage}[b]{\linewidth}\raggedleft
Qwen 3
\end{minipage} & \begin{minipage}[b]{\linewidth}\raggedleft
\end{minipage} \\
\midrule\noalign{}
\endhead
\bottomrule\noalign{}
\endlastfoot
English & eng & Latin & 1.242 & 1.222 & 1.225 & 1.229 & 1.231 & 1.226 &
1.261 & 1.232 & 1.218 & 1.218 & 1.252 & \\
French & fra & Latin & 1.293 & 1.419 & 1.636 & 1.495 & 1.702 & 1.446 &
1.437 & 1.710 & 1.439 & 1.439 & 1.709 & \\
Yoruba & yor & Latin & 1.694 & 2.755 & 3.003 & 2.548 & 2.894 & 2.641 &
3.009 & 3.144 & 2.258 & 2.258 & 2.931 & \\
Hausa & hau & Latin & 1.911 & 1.964 & 2.074 & 1.851 & 2.115 & 1.845 &
2.085 & 2.140 & 1.645 & 1.645 & 2.121 & \\
Igbo & ibo & Latin & 1.822 & 2.514 & 2.534 & 2.274 & 2.525 & 2.142 &
2.574 & 2.594 & 1.731 & 1.731 & 2.551 & \\
Wolof & wol & Latin & 1.834 & 1.942 & 2.019 & 1.909 & 2.048 & 1.900 &
2.058 & 2.084 & 1.847 & 1.847 & 2.090 & \\
Swahili & swh & Latin & 1.600 & 2.293 & 2.460 & 2.083 & 2.460 & 2.121 &
2.370 & 2.490 & 1.872 & 1.872 & 2.507 & \\
Amharic & amh & Ethiopic & 7.912 & 9.839 & 9.329 & 3.035 & 11.905 &
3.738 & 11.936 & 11.921 & 8.969 & 8.969 & 6.453 & \\
Zulu & zul & Latin & 2.864 & 3.297 & 3.520 & 3.183 & 3.533 & 3.158 &
3.476 & 3.576 & 2.749 & 2.749 & 3.593 & \\
Xhosa & xho & Latin & 2.789 & 3.209 & 3.391 & 3.141 & 3.393 & 3.067 &
3.370 & 3.426 & 2.698 & 2.698 & 3.445 & \\
Shona & sna & Latin & 2.836 & 3.027 & 3.240 & 2.915 & 3.239 & 2.908 &
3.171 & 3.329 & 2.622 & 2.622 & 3.350 & \\
Kinyarwanda & kin & Latin & 2.145 & 2.715 & 2.858 & 2.651 & 2.856 &
2.670 & 2.803 & 2.881 & 2.289 & 2.289 & 2.897 & \\
Luganda & lug & Latin & 2.586 & 2.862 & 2.968 & 2.790 & 2.972 & 2.798 &
2.936 & 3.010 & 2.520 & 2.520 & 3.030 & \\
Akan/Twi & aka & Latin & 1.891 & 2.194 & 2.239 & 2.016 & 2.590 & 2.134 &
2.544 & 2.599 & 1.907 & 1.907 & 2.261 & \\
Lingala & lin & Latin & 1.882 & 2.004 & 2.096 & 1.943 & 2.100 & 2.000 &
2.072 & 2.113 & 1.805 & 1.805 & 2.131 & \\
Oromo & gaz & Latin & 3.031 & 2.978 & 3.181 & 2.920 & 3.167 & 2.955 &
3.126 & 3.188 & 2.559 & 2.559 & 3.208 & \\
Sesotho & sot & Latin & 1.848 & 1.968 & 2.043 & 1.880 & 2.051 & 1.871 &
2.042 & 2.073 & 1.700 & 1.700 & 2.088 & \\
Bambara & bam & Latin & 2.277 & 2.535 & 2.610 & 2.490 & 3.017 & 2.501 &
2.991 & 3.066 & 2.414 & 2.414 & 2.594 & \\
Tigrinya & tir & Ethiopic & 7.249 & 9.015 & 8.540 & 3.464 & 10.901 &
4.192 & 10.920 & 10.910 & 8.274 & 8.274 & 5.920 & \\
N'Ko & nqo & N'Ko & 10.866 & 10.864 & 10.865 & 10.725 & 10.865 & 10.865
& 10.865 & 10.869 & 10.865 & 10.865 & 7.465 & \\
Afrikaans & afr & Latin & 1.986 & 1.738 & 1.897 & 1.772 & 1.963 & 1.738
& 1.816 & 1.972 & 1.649 & 1.649 & 1.990 & \\
\end{longtable}
}

\end{landscape}\clearpage\pagestyle{plain}

\begin{landscape}\pagestyle{empty}

\subsubsection{Table B.2: Premium P(L,T) = F(L,T) / F(eng,T) (English =
N/A)}\label{tbl:B.2}

{\def\LTcaptype{none} % do not increment counter
\begin{longtable}[]{@{}
  >{\raggedright\arraybackslash}p{(\linewidth - 28\tabcolsep) * \real{0.1404}}
  >{\raggedright\arraybackslash}p{(\linewidth - 28\tabcolsep) * \real{0.0526}}
  >{\raggedright\arraybackslash}p{(\linewidth - 28\tabcolsep) * \real{0.0702}}
  >{\raggedleft\arraybackslash}p{(\linewidth - 28\tabcolsep) * \real{0.0614}}
  >{\raggedleft\arraybackslash}p{(\linewidth - 28\tabcolsep) * \real{0.0614}}
  >{\raggedleft\arraybackslash}p{(\linewidth - 28\tabcolsep) * \real{0.0614}}
  >{\raggedleft\arraybackslash}p{(\linewidth - 28\tabcolsep) * \real{0.0614}}
  >{\raggedleft\arraybackslash}p{(\linewidth - 28\tabcolsep) * \real{0.0614}}
  >{\raggedleft\arraybackslash}p{(\linewidth - 28\tabcolsep) * \real{0.0614}}
  >{\raggedleft\arraybackslash}p{(\linewidth - 28\tabcolsep) * \real{0.0614}}
  >{\raggedleft\arraybackslash}p{(\linewidth - 28\tabcolsep) * \real{0.0614}}
  >{\raggedleft\arraybackslash}p{(\linewidth - 28\tabcolsep) * \real{0.0614}}
  >{\raggedleft\arraybackslash}p{(\linewidth - 28\tabcolsep) * \real{0.0614}}
  >{\raggedleft\arraybackslash}p{(\linewidth - 28\tabcolsep) * \real{0.0614}}
  >{\raggedleft\arraybackslash}p{(\linewidth - 28\tabcolsep) * \real{0.0614}}@{}}
\toprule\noalign{}
\begin{minipage}[b]{\linewidth}\raggedright
Language
\end{minipage} & \begin{minipage}[b]{\linewidth}\raggedright
ISO
\end{minipage} & \begin{minipage}[b]{\linewidth}\raggedright
Script
\end{minipage} & \begin{minipage}[b]{\linewidth}\raggedleft
BLOOM
\end{minipage} & \begin{minipage}[b]{\linewidth}\raggedleft
Aya
\end{minipage} & \begin{minipage}[b]{\linewidth}\raggedleft
DeepSeek
\end{minipage} & \begin{minipage}[b]{\linewidth}\raggedleft
Gemma 4
\end{minipage} & \begin{minipage}[b]{\linewidth}\raggedleft
Llama 3.1
\end{minipage} & \begin{minipage}[b]{\linewidth}\raggedleft
Llama 4
\end{minipage} & \begin{minipage}[b]{\linewidth}\raggedleft
Tekken
\end{minipage} & \begin{minipage}[b]{\linewidth}\raggedleft
cl100k
\end{minipage} & \begin{minipage}[b]{\linewidth}\raggedleft
o200k
\end{minipage} & \begin{minipage}[b]{\linewidth}\raggedleft
o200k-h
\end{minipage} & \begin{minipage}[b]{\linewidth}\raggedleft
Qwen 3
\end{minipage} & \begin{minipage}[b]{\linewidth}\raggedleft
\end{minipage} \\
\midrule\noalign{}
\endhead
\bottomrule\noalign{}
\endlastfoot
English & eng & Latin & --- & --- & --- & --- & --- & --- & --- & --- &
--- & --- & --- & \\
French & fra & Latin & 1.041 & 1.162 & 1.335 & 1.217 & 1.382 & 1.180 &
1.140 & 1.388 & 1.182 & 1.182 & 1.365 & \\
Yoruba & yor & Latin & 1.364 & 2.256 & 2.451 & 2.073 & 2.351 & 2.154 &
2.386 & 2.552 & 1.854 & 1.854 & 2.342 & \\
Hausa & hau & Latin & 1.538 & 1.608 & 1.693 & 1.507 & 1.718 & 1.505 &
1.653 & 1.738 & 1.351 & 1.351 & 1.694 & \\
Igbo & ibo & Latin & 1.467 & 2.058 & 2.068 & 1.851 & 2.051 & 1.748 &
2.041 & 2.106 & 1.422 & 1.422 & 2.038 & \\
Wolof & wol & Latin & 1.476 & 1.590 & 1.648 & 1.554 & 1.663 & 1.550 &
1.632 & 1.691 & 1.517 & 1.517 & 1.670 & \\
Swahili & swh & Latin & 1.288 & 1.877 & 2.008 & 1.695 & 1.998 & 1.731 &
1.880 & 2.022 & 1.537 & 1.537 & 2.003 & \\
Amharic & amh & Ethiopic & 6.368 & 8.055 & 7.613 & 2.470 & 9.668 & 3.050
& 9.466 & 9.677 & 7.365 & 7.365 & 5.156 & \\
Zulu & zul & Latin & 2.305 & 2.699 & 2.873 & 2.591 & 2.869 & 2.576 &
2.757 & 2.903 & 2.257 & 2.257 & 2.870 & \\
Xhosa & xho & Latin & 2.245 & 2.627 & 2.767 & 2.556 & 2.755 & 2.502 &
2.672 & 2.781 & 2.216 & 2.216 & 2.753 & \\
Shona & sna & Latin & 2.283 & 2.478 & 2.644 & 2.372 & 2.631 & 2.373 &
2.514 & 2.703 & 2.153 & 2.153 & 2.676 & \\
Kinyarwanda & kin & Latin & 1.727 & 2.223 & 2.332 & 2.158 & 2.320 &
2.178 & 2.223 & 2.339 & 1.879 & 1.879 & 2.315 & \\
Luganda & lug & Latin & 2.081 & 2.343 & 2.422 & 2.270 & 2.414 & 2.283 &
2.328 & 2.444 & 2.069 & 2.069 & 2.420 & \\
Akan/Twi & aka & Latin & 1.522 & 1.796 & 1.827 & 1.641 & 2.103 & 1.741 &
2.018 & 2.110 & 1.566 & 1.566 & 1.807 & \\
Lingala & lin & Latin & 1.515 & 1.640 & 1.710 & 1.581 & 1.706 & 1.632 &
1.643 & 1.715 & 1.482 & 1.482 & 1.702 & \\
Oromo & gaz & Latin & 2.440 & 2.438 & 2.596 & 2.377 & 2.572 & 2.410 &
2.479 & 2.588 & 2.101 & 2.101 & 2.563 & \\
Sesotho & sot & Latin & 1.488 & 1.611 & 1.667 & 1.530 & 1.666 & 1.526 &
1.619 & 1.683 & 1.396 & 1.396 & 1.668 & \\
Bambara & bam & Latin & 1.833 & 2.076 & 2.130 & 2.026 & 2.450 & 2.041 &
2.372 & 2.489 & 1.982 & 1.982 & 2.072 & \\
Tigrinya & tir & Ethiopic & 5.835 & 7.380 & 6.969 & 2.819 & 8.853 &
3.420 & 8.659 & 8.857 & 6.794 & 6.794 & 4.729 & \\
N'Ko & nqo & N'Ko & 8.746 & 8.894 & 8.866 & 8.729 & 8.824 & 8.864 &
8.616 & 8.824 & 8.922 & 8.922 & \textbf{5.964} & \\
Afrikaans & afr & Latin & 1.598 & 1.423 & 1.548 & 1.442 & 1.594 & 1.418
& 1.440 & 1.601 & 1.354 & 1.354 & 1.590 & \\
\end{longtable}
}

\end{landscape}\clearpage\pagestyle{plain}

\subsection{Appendix C: Reproducibility
Manifest}\label{appendix-c-reproducibility-manifest}

All artifacts required to reproduce the headline results from scratch
are listed below. The \texttt{runs/main/manifest.json} file records the
exact run configuration.

\subsubsection{C.1 Software Versions (Main Run,
2026-06-15)}\label{c.1-software-versions-main-run-2026-06-15}

\phantomsection\label{tbl:C.1}

\textbf{Table C.1: Software versions, main run (2026-06-15)}

{\def\LTcaptype{none} % do not increment counter
\begin{longtable}[]{@{}lll@{}}
\toprule\noalign{}
Component & Version & Notes \\
\midrule\noalign{}
\endhead
\bottomrule\noalign{}
\endlastfoot
\texttt{afri-fertility} & 0.1.0 & This paper's measurement tool \\
Python & 3.11.15 & conda env \texttt{afri-fertility} \\
\texttt{transformers} & 5.12.x & HuggingFace tokenizers backend \\
\texttt{tokenizers} & 0.22.x & Fast tokenizer runtime \\
\texttt{tiktoken} & ≥0.7 & OpenAI tokenizer backend \\
\texttt{datasets} & ≥2.19 & FLORES-200+, SIB-200, MAFAND-MT loading \\
\texttt{numpy} & ≥1.26 & Bootstrap CI computation \\
\texttt{pandas} & ≥2.2 & Results aggregation \\
\end{longtable}
}

Exact pinned versions of all HuggingFace model snapshots are recorded in
the HuggingFace Hub cache manifest. Tokenizer IDs and HuggingFace repo
names are listed in
\texttt{src/afri\_fertility/tokenizers/hf\_adapter.py}.

\subsubsection{C.2 Run Configuration}\label{c.2-run-configuration}

\phantomsection\label{tbl:C.2}

\textbf{Table C.2: Run configuration parameters}

{\def\LTcaptype{none} % do not increment counter
\begin{longtable}[]{@{}
  >{\raggedright\arraybackslash}p{(\linewidth - 2\tabcolsep) * \real{0.5000}}
  >{\raggedright\arraybackslash}p{(\linewidth - 2\tabcolsep) * \real{0.5000}}@{}}
\toprule\noalign{}
\begin{minipage}[b]{\linewidth}\raggedright
Parameter
\end{minipage} & \begin{minipage}[b]{\linewidth}\raggedright
Value
\end{minipage} \\
\midrule\noalign{}
\endhead
\bottomrule\noalign{}
\endlastfoot
Unicode normalization & NFC \\
Word segmentation & UAX-29 (uniseg) \\
Aggregation method & sum-then-divide (sum tokens / sum words across all
sentences) \\
Bootstrap iterations & 1,000 \\
Bootstrap seed & 42 \\
Baseline language & English (\texttt{eng}) \\
Primary corpus & FLORES-200+ devtest \\
Config file & \texttt{configs/study\_main.yaml} \\
Run timestamp & 2026-06-15T20:54:49 UTC \\
Records produced & 616 (22 languages × 3 corpora × 11 tokenizers, with
some corpus/language gaps) \\
Active tokenizers & 11 (qwen/qwen3 re-run after cache fix; all data
valid; see Deviation A.2) \\
\end{longtable}
}

\subsubsection{C.3 Price and FX
Snapshots}\label{c.3-price-and-fx-snapshots}

\phantomsection\label{tbl:C.3}

\textbf{Table C.3: API price and FX rate snapshot files}

{\def\LTcaptype{none} % do not increment counter
\begin{longtable}[]{@{}lll@{}}
\toprule\noalign{}
Snapshot & File & Date \\
\midrule\noalign{}
\endhead
\bottomrule\noalign{}
\endlastfoot
API token prices & \texttt{configs/prices\_2026-06.yaml} & 2026-06-12 \\
FX rates & \texttt{configs/fx\_2026-06.yaml} & 2026-06-12 \\
\end{longtable}
}

All prices are in USD per token. FX rates: 1 USD = ₦1,360.95 (NGN) /
R16.31 (ZAR) / Ksh129.45 (KES) (snapshot date 2026-06-12). Sources and
verification dates are annotated inline in the price YAML file
(\texttt{configs/prices\_2026-06.yaml}).

\subsubsection{C.4 Released Artifacts}\label{c.4-released-artifacts}

\phantomsection\label{tbl:C.4}

\textbf{Table C.4: Released artifacts and access locations}

{\def\LTcaptype{none} % do not increment counter
\begin{longtable}[]{@{}
  >{\raggedright\arraybackslash}p{(\linewidth - 4\tabcolsep) * \real{0.3333}}
  >{\raggedright\arraybackslash}p{(\linewidth - 4\tabcolsep) * \real{0.3333}}
  >{\raggedright\arraybackslash}p{(\linewidth - 4\tabcolsep) * \real{0.3333}}@{}}
\toprule\noalign{}
\begin{minipage}[b]{\linewidth}\raggedright
Artifact
\end{minipage} & \begin{minipage}[b]{\linewidth}\raggedright
Location
\end{minipage} & \begin{minipage}[b]{\linewidth}\raggedright
License
\end{minipage} \\
\midrule\noalign{}
\endhead
\bottomrule\noalign{}
\endlastfoot
\texttt{afri-fertility} tool & PyPI: \texttt{afri-fertility} &
Apache-2.0 \\
Results dataset & Hugging Face:
\texttt{datalens-africa/afri-fertility-results} & CC-BY-4.0 \\
Leaderboard JSON & \texttt{papers/p1-token-tax/leaderboard.json} in this
repo & CC-BY-4.0 \\
Study config + price/FX snapshots & \texttt{configs/} in
\texttt{afri-fertility} repo & Apache-2.0 \\
\end{longtable}
}

\subsubsection{C.5 Reproduction Steps}\label{c.5-reproduction-steps}

\begin{Shaded}
\begin{Highlighting}[]
\FunctionTok{git}\NormalTok{ clone https://github.com/CipherSenseAI/afri{-}fertility}
\BuiltInTok{cd}\NormalTok{ afri{-}fertility}
\ExtensionTok{pip}\NormalTok{ install }\AttributeTok{{-}e} \StringTok{".[dev,viz]"}

\CommentTok{\# Offline credibility check (no API keys, no HF downloads needed for tiktoken)}
\ExtensionTok{afri{-}fertility}\NormalTok{ reproduce}

\CommentTok{\# Full locked study (requires HF\_TOKEN for gated tokenizers)}
\BuiltInTok{export} \VariableTok{HF\_TOKEN}\OperatorTok{=}\NormalTok{hf\_...}
\ExtensionTok{afri{-}fertility}\NormalTok{ run }\AttributeTok{{-}{-}config}\NormalTok{ configs/study\_main.yaml}

\CommentTok{\# Regenerate figures}
\ExtensionTok{afri{-}fertility}\NormalTok{ figures }\AttributeTok{{-}{-}run}\NormalTok{ runs/main}

\CommentTok{\# Emit leaderboard JSON}
\ExtensionTok{afri{-}fertility}\NormalTok{ leaderboard }\AttributeTok{{-}{-}run}\NormalTok{ runs/main }\AttributeTok{{-}{-}out}\NormalTok{ leaderboard.json}
\end{Highlighting}
\end{Shaded}

All tokenizer versions, price snapshot date, FX rates, and a SHA-256
config hash are recorded in \texttt{runs/main/manifest.json}.

\section*{References}\label{references}

\protect\phantomsection\label{ref:ahia2023}{}Ahia, O., Kumar, S., Gonen,
H., Kasai, J., Mortensen, D. R., Smith, N. A., \& Tsvetkov, Y. (2023).
Do All Languages Cost the Same? Tokenization in the Era of Commercial
Language Models. In \emph{Proceedings of the 2023 Conference on
Empirical Methods in Natural Language Processing} (pp.~9904--9923).
Association for Computational Linguistics.
\url{https://aclanthology.org/2023.emnlp-main.614/} (arXiv:
\url{https://arxiv.org/abs/2305.13707})

\protect\phantomsection\label{ref:adelani2025iroko}{}Adelani, D. I.,
Ojo, J., Azime, I. A., Zhuang, J. Y., Alabi, J. O., He, X., Ochieng, M.,
Hooker, S., Bukula, A., Lee, E.-S. A., Chukwuneke, C. I., Buzaaba, H.,
Sibanda, B. K., Kalipe, G. K., Mukiibi, J., Kabongo Kabenamualu, S.,
Yuehgoh, F., Setaka, M., Ndolela, L., Odu, N., Mabuya, R., Muhammad, S.
H., Osei, S., Samb, S., Guge, T. K., \& Stenetorp, P. (2025).
IrokoBench: A New Benchmark for African Languages in the Age of Large
Language Models. In \emph{Proceedings of the 2025 Conference of the
North American Chapter of the Association for Computational Linguistics:
Human Language Technologies (Volume 1: Long Papers)} (pp.~2732--2757).
Association for Computational Linguistics.
\url{https://doi.org/10.18653/v1/2025.naacl-long.139} (arXiv:
\url{https://arxiv.org/abs/2406.03368})

\protect\phantomsection\label{ref:gebru2021}{}Gebru, T., Morgenstern,
J., Vecchione, B., Vaughan, J. W., Wallach, H., Daumé III, H., \&
Crawford, K. (2021). Datasheets for Datasets. \emph{Communications of
the ACM}, 64(12), 86--92. \url{https://doi.org/10.1145/3458723}

\protect\phantomsection\label{ref:nllb2024}{}NLLB Team. (2024). Scaling
neural machine translation to 200 languages. \emph{Nature}, 630(8018),
841--846. \url{https://doi.org/10.1038/s41586-024-07335-x}

\protect\phantomsection\label{ref:ndomba2025}{}Ndomba, G. E., Mswahili,
M. E., \& Jeong, Y.-S. (2025). Tokenizers for African Languages.
\emph{IEEE Access}, 13, 1046--1054.
\url{https://doi.org/10.1109/ACCESS.2024.3522285}

\protect\phantomsection\label{ref:ojo2023}{}Ojo, J., Ogundepo, O.,
Oladipo, A., Ogueji, K., Lin, J., Stenetorp, P., \& Adelani, D. I.
(2023). AfroBench: How Good are Large Language Models on African
Languages? arXiv preprint arXiv:2311.07978.
\url{https://arxiv.org/abs/2311.07978}

\protect\phantomsection\label{ref:ovcharov2026}{}Ovcharov, V. (2026).
The Tokenizer Tax Across 25 European Languages: Domain Invariance,
Cross-Lingual Few-Shot Effects, and the Ukrainian Penalty. arXiv
preprint arXiv:2605.24718. \url{https://arxiv.org/abs/2605.24718}

\protect\phantomsection\label{ref:petrov2023}{}Petrov, A., La Malfa, E.,
Torr, P., \& Bibi, A. (2023). Language Model Tokenizers Introduce
Unfairness Between Languages. In \emph{Advances in Neural Information
Processing Systems} (NeurIPS 2023), Vol. 36 (pp.~36963--36990).
\url{https://arxiv.org/abs/2305.15425}

\protect\phantomsection\label{ref:rust2021}{}Rust, P., Pfeiffer, J.,
Vulić, I., Ruder, S., \& Gurevych, I. (2021). How Good is Your
Tokenizer? On the Monolingual Performance of Multilingual Language
Models. In \emph{Proceedings of the 59th Annual Meeting of the
Association for Computational Linguistics and the 11th International
Joint Conference on Natural Language Processing} (ACL-IJCNLP 2021), Vol.
1 (Long Papers) (pp.~3118--3135). Association for Computational
Linguistics. \url{https://doi.org/10.18653/v1/2021.acl-long.243}

\protect\phantomsection\label{ref:lundin2026}{}Lundin, J. M., Zhang, A.,
Karim, N., Louzan, H., Wei, V., Adelani, D. I., \& Carroll, C. (2026).
The Token Tax: Systematic Bias in Multilingual Tokenization. In
\emph{Proceedings of the 7th Workshop on African Natural Language
Processing (AfricaNLP 2026)} (pp.~103--112). Association for
Computational Linguistics.
\url{https://doi.org/10.18653/v1/2026.africanlp-main.10} (arXiv:
\url{https://arxiv.org/abs/2509.05486})

\subsection*{Corpora and datasets}\label{corpora-and-datasets}

\protect\phantomsection\label{ref:mafand2022}{}Adelani, D. I., Alabi, J.
O., Fan, A., Kreutzer, J., Shen, X., Reid, M., Ruiter, D., Klakow, D.,
Nabende, P., Chang, E., Gwadabe, T. G., Sackey, F., Dossou, B. F. P.,
Emezue, C. C., Leong, C., Beukman, M., Muhammad, S. H., Jarso, G. D.,
Yousuf, O., Rubungo, A. N. N., Hacheme, G., Wairagala, E. P., Nasir, M.
U., Ajibade, B. A., Ajayi, T. O., Gitau, Y. W., Abbott, J., Ahmed, M.,
Ochieng, M., Aremu, A., Ogayo, P., Mukiibi, J., Kabore, F. O., Kalipe,
G. K., Mbaye, D., Tapo, A. A., Koagne, V. M., Munkoh-Buabeng, E.,
Wagner, V., Abdulmumin, I., Awokoya, A., Buzaaba, H., Sibanda, B. K.,
Bukula, A., \& Manthalu, S. (2022). A Few Thousand Translations Go a
Long Way! Leveraging Pre-trained Models for African News Translation. In
\emph{Proceedings of the 2022 Conference of the North American Chapter
of the Association for Computational Linguistics: Human Language
Technologies} (NAACL-HLT 2022) (pp.~3053--3070). Association for
Computational Linguistics.
\url{https://aclanthology.org/2022.naacl-main.223}

\protect\phantomsection\label{ref:sib2024}{}Adelani, D. I., Liu, H.,
Shen, X., Vassilyev, N., Alabi, J. O., Mao, Y., Gao, H., \& Lee, E.-S.
A. (2024). SIB-200: A Simple, Inclusive, and Big Evaluation Dataset for
Topic Classification in 200+ Languages and Dialects. In
\emph{Proceedings of the 18th Conference of the European Chapter of the
Association for Computational Linguistics (Volume 1: Long Papers)}
(pp.~226--245). Association for Computational Linguistics.
\url{https://aclanthology.org/2024.eacl-long.14} (arXiv:
\url{https://arxiv.org/abs/2309.07445})

\end{document}